\documentclass[sigconf]{acmart}

\usepackage{booktabs} 
\usepackage{helvet}
\usepackage{courier}
\usepackage{amsmath}
\usepackage{amsfonts}
\usepackage{algorithm}
\usepackage[noend]{algorithmic}
\usepackage{makecell,multirow,diagbox}
\usepackage{mathrsfs}
\usepackage{graphicx}
\usepackage{subfigure}
\usepackage{comment}
\usepackage{booktabs}
\usepackage{caption}
\usepackage{float}
\usepackage{color}
\usepackage{footnote}

\setcopyright{none}
\renewcommand\footnotetextcopyrightpermission[1]{}
\def\BibTeX{{\rm B\kern-.05em{\sc i\kern-.025em b}\kern-.08emT\kern-.1667em\lower.7ex\hbox{E}\kern-.125emX}}

\thanks{This paper has been accepted by CIKM2019. \\The previous name is: Learning to Advertise with Adaptive Exposure via Constrained Two-Level Reinforcement Learning}

\author{Weixun Wang}
\email{wxwang@tju.edu.cn}
\affiliation{%
  \institution{Tianjin University}
}

\author{Junqi Jin}
\email{junqi.jjq@alibaba-inc.com}
\affiliation{%
  \institution{Alibaba Group}
}

\author{Jianye	Hao}
\email{jianye.hao@tju.edu.cn}
\affiliation{%
  \institution{Tianjin University}
}

\author{Chunjie	Chen, Chuan	Yu}
\email{{chunjie.ccj,yuchuan.yc}@alibaba-inc.com}
\affiliation{%
  \institution{Alibaba Group}
}

\author{Weinan	Zhang}
\email{wnzhang@sjtu.edu.cn}
\affiliation{%
  \institution{Shanghai Jiao Tong University	}
}

\author{Jun	Wang}
\email{jun.wang@cs.ucl.ac.uk}
\affiliation{%
  \institution{University College London}
}

\author{Xiaotian Hao}
\email{xiaotianhao@tju.edu.cn}
\affiliation{%
 \institution{Tianjin University}
}

\author{Yixi Wang}
\email{yixiwang2017@outlook.com}
\affiliation{%
 \institution{Tianjin University}
}

\author{Han Li, Jian Xu, Kun Gai}
\email{{lihan.lh, xiyu.xj, jingshi.gk}@alibaba-inc.com}
\affiliation{%
 \institution{Alibaba Group}
}

\setcopyright{acmlicensed}
\begin{document}

%
\title{Learning Adaptive Display Exposure \\ for Real-Time Advertising}

\begin{abstract}
In E-commerce advertising, where product recommendations and product ads are presented to users simultaneously, the traditional setting is to display ads at fixed positions. However, under such a setting, the advertising system loses the flexibility to control the number and positions of ads, resulting in sub-optimal platform revenue and user experience. Consequently, major e-commerce platforms (e.g., Taobao.com) have begun to consider more flexible ways to display ads. In this paper, we investigate the problem of \emph{advertising with adaptive exposure:} can we dynamically determine the number and positions of ads for each user visit under certain business constraints so that the platform revenue can be increased? More specifically, we consider two types of constraints: \emph{request-level} constraint ensures user experience for each user visit, and \emph{platform-level} constraint controls the overall platform monetization rate. We model this problem as a Constrained Markov Decision Process with per-state constraint (psCMDP) and propose a constrained two-level reinforcement learning approach to decompose the original problem into two relatively independent sub-problems. To accelerate policy learning, we also devise a constrained hindsight experience replay mechanism. Experimental evaluations on industry-scale real-world datasets demonstrate the merits of our approach in both obtaining higher revenue under the constraints and the effectiveness of the constrained hindsight experience replay mechanism.
\end{abstract}

%
%


%

%

%
\maketitle

\begin{figure*}[t]
\centering
\includegraphics[width=6.0 in,angle=0]{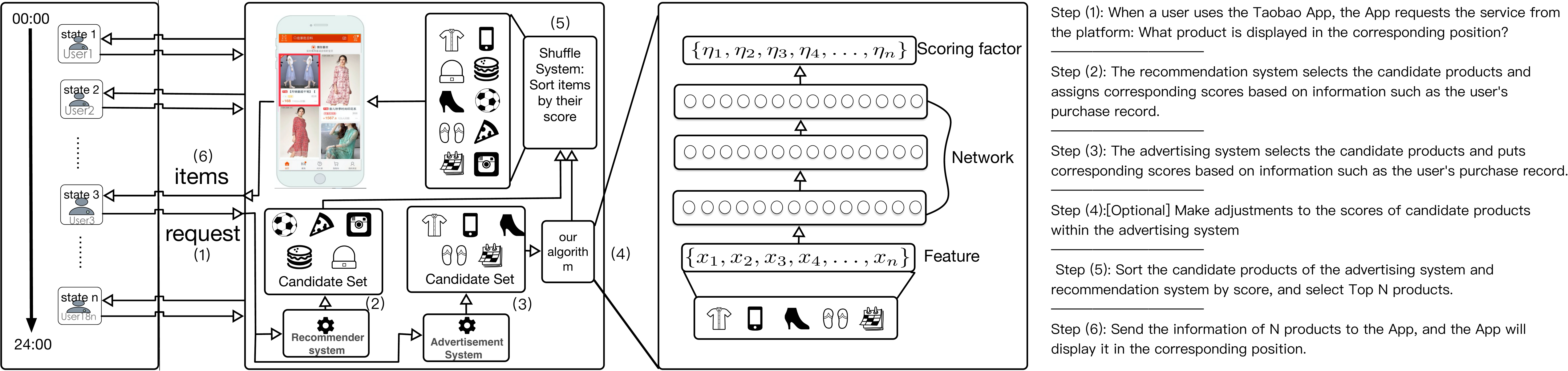}
\caption{Advertising with Adaptive Exposure and Our System Structure.}
\label{fig_ill}
\end{figure*}

\section{Introduction}
With the advances of deep neural network \cite{lecun2015deep,goodfellow2016deep}, Deep Reinforcement Learning (DRL) approaches have made significant progress in a number of applications including Atari games \cite{mnih2015human} and robot locomotion and manipulation \cite{schulman2015high,levine2016end}. Recently, we also witness successful applications of DRL techniques to optimize the decision-making process in E-commerce from different aspects including online recommendation \cite{chen2018stabilizing}, impression allocation \cite{cai2018reinforcement1,zhao2018impression}, advertising bidding strategies \cite{jin2018real,wu2018budget,zhao2018deep} and product ranking \cite{hu2018reinforcement}. 

In traditional online advertising, the ad positions are fixed, and we only need to determine which ads to be shown in these positions for each user request \cite{mehta2013online}. This can be modeled as an ads position bidding problem and DRL techniques have been shown to be effective in learning bidding strategies for advertisers \cite{jin2018real,wu2018budget,zhao2018deep}. However, fixing ad positions limit the flexibility of the advertising system. Intuitively, if a user is with high monetization value (e.g., likes to click ads), it is reasonable for the advertising platform to display more ads when this user visits. On the other hand, we are also concerned with displaying too many ads for two reasons. First, it might lead to poor user experience and have a negative impact on user retention. Second, monetization rate is an important business index for a company to moderate. Therefore, in this paper, we consider two levels of constraints: (1) \emph{request-level}: the maximum number of ads on each request \footnote{It is worth pointing out that the request here refers to the user's access to the platform (for example: opening the mobile app, swipe the screen)} (a.k.a. user visit) cannot exceed 
a threshold; and (2) \emph{platform-level}: the average number of ads over all the requests (within a time window) cannot exceed
a threshold. Under the above constraints, we investigate the problem of advertising with adaptive exposure: can we dynamically determine the set of ads and their positions for each user visit so that the platform revenue can be maximized? We call the above problem as \textbf{\textit{advertising with adaptive exposure problem}}.

Fig.1 illustrates the flexible adaptive exposure mechanism adopted by Taobao \footnote{One of the largest} e-commerce company in China. For each user visit, the platform presents a dynamic mixture of product recommendations and product ads. The ad positions are not a fixed prior, and they are determined by the user's profile and behaviors. The adaptive exposure problem can be formalized as a sequential decision problem. In each step, the recommendation and the advertising systems first select some items based on their scoring systems independently. Then these commodities are sorted altogether by their scores and the top few items are exposed to the request (user). 

We model the above problem as a Constrained Markov Decision Process (CMDP) \cite{altman1999constrained}. Although optimal policies for small-sized CMDPs can be derived using linear programming \cite{altman1999constrained}, it is difficult to construct such policies for  large-scale and complex real-world e-commerce platforms. Thus, we resort to model-free RL approaches to learn approximately optimal solutions \cite{achiam2017constrained,tessler2018reward}. Existing model-free RL approaches for solving CMDP are trajectory-based: they update policies by propagating constraint-violation signals over the entire trajectory \cite{achiam2017constrained,prashanth2016variance}. Unfortunately, most of them fail to meet the constraints \cite{tessler2018reward}. To address this issue, \citet{tessler2018reward} propose the Reward Constrained Policy Optimization (RCPO), which decomposes the trajectory constraints into per-state penalties and dynamically adjusts their weights. To ensure that the overall penalty of a trajectory satisfies the given constraint, the constraint-violation signals are also propagated back along the entire trajectory. However, in the advertising with adaptive exposure problem, we need to satisfy both state-level (request-level) and trajectory-level (platform-level) constraints. RCPO only considers trajectory-level constraints and thus cannot be directly applied here. 

In this paper, we first model the \textit{advertising with adaptive exposure problem} as a CMDP with per-state constraint (psCMDP). Then we propose a constrained two-level reinforcement learning framework to learn optimal advertising policies satisfying both state-level and trajectory-level constraints. In our framework, the trajectory-level constraint and the state-level constraint are divided into different levels in the learning process. The higher level policy breaks a trajectory into multiple sub-trajectories and tackles the problem of selecting constraints for each sub-trajectory to maximize total revenue under the trajectory-level constraint. Each sub-trajectory identifies an independent optimization problem with both sub-trajectory constraint and state-level constraint. Here we simplify the sub-trajectory optimization problem at the cost of sacrificing the policy optimality by treating the sub-trajectory constraint as another state-level constraint. In this way, we can easily combine the sub-trajectory constraint with the original state-level constraint and use off-policy methods such as Deep Deterministic Policy Gradient (DDPG) \cite{lillicrap2015continuous} with auxiliary task \cite{jaderberg2016reinforcement} to train the lower level policy. We also propose Constrained Hindsight Experience Replay (CHER) to accelerate the lower level policy training.

Note that our framework can be naturally extended to more levels by further decomposing each sub-trajectory into a number of sub-trajectories. Thus it is expected that the quality of the learned policy would be improved when we increase the number of levels, which means the length of each sub-trajectory at the lower levels is reduced. Thus our framework is flexible enough to make a compromise between training efficiency and policy optimality. In this paper, we set our framework to be two levels. One additional benefit of our two-level framework is that we can easily reuse the lower level policy to train the higher level constraint selection policy in case the trajectory-level constraint is adjusted. We evaluate our approach using real-world datasets from Taobao platform both offline and online. Our approach can improve the advertising revenue and the advertisers' income while satisfying the constraints at both levels. In the lower level, we verify  that CHER mechanism can significantly improve the training speed and reduce the deviation of the per-state constraint. Moreover, in the higher level, our method can make good use of the lower level policy set to learn higher level policies with respect to different platform-level constraints.


\section{Preliminary: Constrained Reinforcement Learning}
Reinforcement learning (RL) allows agents to interact with the environment by sequentially taking actions and observing rewards to maximize the cumulative reward \cite{sutton1998reinforcement}. RL can be modeled as a Markov Decision Process (MDP), which is defined as a tuple $(S,A,R,P)$. $S$ is the state space and $A$ is the action space. The immediate reward function is $R: S \times A \times S \rightarrow \mathbb{R}$. $P$ is the state transition probability, $S \times A \times S \rightarrow [0,1]$. There exists a policy $\pi$ on $A$, which defines an agent's behavior. The agent uses its policy to interact with the environment and generates a trajectory $\tau:\ \{s_0, a_0, r_0, s_1, \dots s_{t}, a_{t}, r_{t}, s_{t+1},\dots\}$. Its goal is to learn an optimal policy $\pi^*$ which maximizes the expected return given the initial state:
\begin{equation}\label{eq_op_crl}
\pi^*= \underset{\pi}{\arg\max}\ E[\sum_{t=0}^{T}\gamma^{t} r_t|\pi]
\tag{1} \end{equation}
Here $\gamma \in [0,1]$ is the discount factor, $T$ is the length of the trajectory $\tau$. The Constrained Markov Decision Process (CMDP) \cite{altman1999constrained} is generally used to deal with the situation, by which the feasible policies are restricted. Specifically, CMDP is augmented with auxiliary cost functions $C_T$, $S \times A \times S \rightarrow \mathbb{R}$ and a upperbound constraint $u_T$. Let $J_{C_T}(\pi)$ be the cumulative discounted cost of policy $\pi$. The expected discounted return is defined as follows:
\begin{equation}
J_{C_T}(\pi) = \underset{\tau \sim \pi}{E}[\sum_{t=0}^{T}\gamma^{t} C_T(s_t, a_t, s_{t+1})|\pi]
\tag{2} \end{equation}
The set of feasible stationary policies for a CMDP is then:
\begin{equation}
\Pi_C \doteq \{\pi \in \Pi : J_{C_T}(\pi) \le u_T \}
\label{cmdp_p}
\tag{3} \end{equation}
And the policy is optimized by limiting the policy $\pi \in \Pi_C$ in the Equation \ref{eq_op_crl}.
For DRL methods, \citet{achiam2017constrained} propose a new approach which replaces the optimization objective and constraints with surrogate functions, and uses Trust Region Policy Optimization \cite{schulman2015trust} to learn the policy, achieving near-constraint satisfaction in each iteration. \citet{tessler2018reward} use a method similar to WeiMDP \cite{geibel2006reinforcement}. WeiMDP \cite{geibel2006reinforcement} introduces a weight parameter $\zeta \in [0, 1]$ and a derived weighted reward function $r'_t$, which is defined as:
\begin{equation}
r'_t = \zeta \times r_t + (1 - \zeta) \times C_T(s_t, a_t, s_{t+1})
\label{add_reward}
\tag{4}
\end{equation}
where $r_t$ and $C_T(s_t, a_t, s_{t+1})$ are the reward and the auxiliary cost under the transition $(s_t, a_t, s_{t+1})$ respectively. For a fixed $\zeta$, this new unconstrained MDP can be solved with standard methods, e.g. Q-Learning \cite{sutton1998reinforcement}. \citet{tessler2018reward} use the weight $\zeta$ as the input to the value function and dynamically adjust the weight by backpropagation.

\begin{table}[t]
\centering
\fontsize{8}{10}\selectfont
\caption{List of notations.}\label{tab1}
\begin{tabular}{l p{6.3cm}}
\toprule
Notation& Description\\
\midrule
$\mathcal{Q}$ & The sequence of incoming requests. $\mathcal{Q}=\{q_1, q_2, ..., q_M\},|\mathcal{Q}|=M$, $M$ is the total number of requests visiting the platform within one day, different days have different $M$.\\
$q_i$& The $i$-th request in the day. \\
$N^{d}$ & the number of candidate ads.\\
$N^{r}$ & The number of recommended products.\\
$N_{i}$ & The number of commodities shown to $q_i$. Usually, $N_{i} = N_{i'}, \forall 1 \le i,i' \le M$. And $N_i < N^{d} + N^{r}$\\
$N_{i}^{d}$ & The number of ads exposed for $q_i$\\
$PV\hspace{-1pt}R$& The total percentage of ads exposed in one day. $PV\hspace{-1pt}R = \frac{\Sigma_{1\le i \le M} N_{i}^{d}}{\Sigma_{1\le i \le M} N_{i}}$\\
$PV\hspace{-1pt}R_i$& The percentage of ads exposed for $q_i$. $PV\hspace{-1pt}R_i = \frac{N_{i}^{d}}{N_{i}}$.\\
$\alpha$ & The maximum percentage of the total ads exposed in one day\\
$\beta$& The maximum percentage of the ads exposed for each request\\
$\mathcal{D}_i$ & The candidate ads set for $q_i$. $|\mathcal{D}_i|=N^{d}$\\
\bottomrule
\end{tabular}
\end{table}

\section{Advertising With Adaptive Exposure}
\subsection{Adaptive Exposure Mechanism}\label{AEM}

In an E-commerce platform, user requests come in order, $\mathcal{Q}=\{q_1, q_2, ..., q_M\}$\footnote{Table 1 summarizes the notations. In this paper, we usually use i subscript to refer to the i-th request and j subscript to refer to the j-th ad in request.}. When a user sends a shopping request $q_i$, $N_{i}$ commodities are exposed to the request based on the user's shopping history and personal preferences. The $N_{i}$ commodities are composed of advertising and recommendation products.  Exposing more ads may increase the advertising revenue. However, the exposed ads for users are not necessarily their favorite or needed products. Therefore, we should limit the number of exposed ads for each user's request.

For each request $q_i$, traditional E-commerce systems use fixed-positions to expose ads: $N^{d}_{i} = K, \forall 1 \le i \le M $, where $K$ is the number of fixed positions.
However, it is obvious that this advertising mechanism is not optimal. Different consumers have different shopping habits and preferences to different products and ads. Therefore, we can expose more advertising products to those consumers who are more likely to click and purchase them (thus increasing the advertising revenue) and vice versa.

To this end, recently Taobao (one of the largest Chinese E-commerce platform)
has begun to adopt more flexible and mixed exposing mechanism for exposing advertising and recommended products (Fig. \ref{fig_ill}). Specifically, for each request $q_i$, the recommendation and the advertising systems first select the top $N^{r}$ and $N^{d}$ items based on their scoring systems independently (Fig. \ref{fig_ill}, Step (2) and Step (3)). Then these commodities are sorted altogether according to scores in descending order(Fig. \ref{fig_ill}, Step (5)) and the top $N_{i}$ items are exposed to this request (Fig. \ref{fig_ill}, Step (6)).

In the meantime, to ensure users' shopping experience, we need to impose the following constraints:
\begin{itemize}
\item \textbf{\emph{platform-level} constraint} $C_{platform}$, the total percentage of ads exposed in one day should not exceed a certain threshold $\alpha$:
\begin{equation}
PV\hspace{-1pt}R \leq \alpha 
\label{con2}
\tag{5} \end{equation}
\item \textbf{\emph{request-level} constraint} $C_{request}$, the percentage of ads exposed for each request should not exceed a certain threshold $\beta$:
\begin{equation}
PV\hspace{-1pt}R_i \leq \beta, \forall 1\le i \le M \label{con1}
\tag{6} \end{equation}
\end{itemize}
where $\alpha \le \beta$. This means that we can exploit this inequality requirement to expose different numbers of ads to different requests according to users' profiles, e.g. expose more ads to users who are more interested in the ads and can increase the average advertising revenue, and fewer ads for the others. On one hand, for each request, the size of $N_{i}^{d}$ can be automatically adjusted according to the quality of the candidate products and ads. On the other hand, the positions of $N_{i}$ items are determined by the quality of the products and ads, which can further optimize the user experience. In this way, we can increase the total advertising revenue while satisfying both the \emph{request-level} and \emph{platform-level} constraints. The scoring systems for both recommendation and advertising sides can be viewed as black boxes. However, from the advertising perspective, we can adaptively adjust the score of each ad to change their relative rankings and the number of ads to be exposed eventually (Fig. \ref{fig_ill}, Step (4)). 

 The adaptive exposure mechanism can potentially improve the advertising revenue, however, it faces a number of challenges. First, the ads score adjusting strategy is highly sensitive to the dynamics of recommendation system (e.g., system upgrading) and other components of advertising system (e.g., the candidate ads selection mechanism may upgraded). Our advertising system needs to be adjusted to meet the constraints. Second, actual business needs to change from time to time (adjust to \emph{platform-level} and \emph{request-level} constrain), so does our advertising system. These challenges force us to design a more flexible algorithm. \footnote{Due to space constraints, we further discuss the novelty and related work of our setup and methods in the appendix.}

\begin{figure*}[t]
\centering
\includegraphics[width=5.5 in,angle=0]{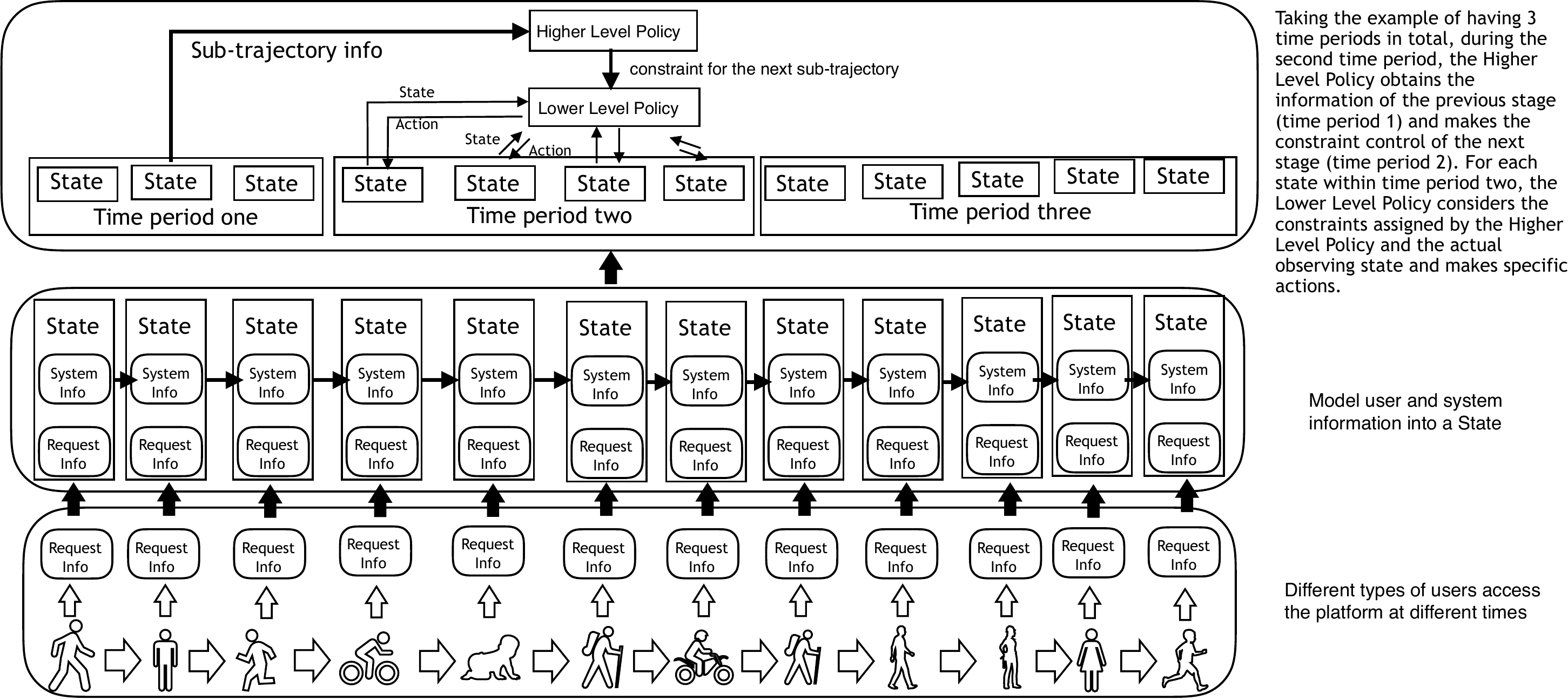}
\caption{The Framework Structure.}
\label{temporal}
\end{figure*}

\subsection{Formulation}
\subsubsection{Problem Description}
From the advertising perspective, the above advertising exposure problem can be seen as a bidding problem:  The product displayed in each user request is determined by the score (bid price) of the advertising item and the recommended item (Rank by bid price, the higher the price, the more possible to be displayed), and the advertising system adjusts the score of the original advertisement (the auction bid) to satisfy the constraint and increase revenue (auction results).
We follow the settings of the bidding problem in Display Advertising \cite{cai2017real,zhang2014optimal} and extend it to \textit{advertising with adaptive exposure problem}.
Formally, for the $j$-th ad $d_j \in \mathcal{D}_i$ in request $q_i$, its score is adjusted as follows:
\begin{equation}
score_{i,j}' = score_{i,j} \times \eta_{i,j}
\tag{7}
\end{equation}
where $\eta_{i,j} = b(q_i, d_j;\theta)$. $b$ is a bidding function and $\theta$ is the parameters of $b$. $score_{i,j}$ is the original score given by the advertising system for $d_j$ in request $q_i$.
Within the advertising system only, we cannot directly figure out whether the ad (which score has been adjusted) will be finally exposed to the request. We can only get the final displayed results from the Shuffle System  (Fig. \ref{fig_ill}, Step (5)).
 So we define $w(b(q_i, d_j;\theta), q_i, d_j) = E_{\phi}[I(b(q_i, d_j;\theta), \phi)]$ as the probability of winning the bid request $(q_i, d_j)$ with bid adjustment ratio $b(q_i, d_j;\theta)$, 
 where $\phi$ is the parameter of the recommendation system and $I(b(q_i, d_j;\theta), \phi)$ indicates whether the advertisement $d_j$ is finally displayed in request $q_i$ given the recommendation system's parameters $\phi$.
We use $v(b(q_i, d_j;\theta), q_i, d_j)$
to denote the expected revenue value of the advertising product $d_j$ under request $q_i$.\footnote{  $v(b(q_i, d_j;\theta), q_i, d_j)$ can be computed in a truthful or Generalized Second Price (GSP) fashion.} Then under the premise of satisfying the constraints $C_{request}$ and $C_{platform}$, the optimization goal of the advertising system can be written as follows:
\begin{align*}
&\underset{\theta}{\max} \sum_{q_i \in \mathcal{Q}}\sum_{d_j \in D_i}  v(b(q_i, d_j;\theta), q_i, d_j) \times w(b(q_i, d_j;\theta), q_i, d_j)\\
&s.t.\quad
\begin{cases}
\frac{\underset{d_j \in D_i}{\sum} w(b(q_i, d_j;\theta), q_i, d_j)}{N_i} \leq \beta, \forall q_i \in \mathcal{Q}\\
\frac{\underset{q_i \in \mathcal{Q}}{\sum}\hspace{2pt} \underset{d_j \in D_i}{\sum} w(b(q_i, d_j;\theta), q_i, d_j)}{\underset{q_i \in \mathcal{Q}}{\sum} N_i} \leq \alpha
\end{cases}
\tag{8} 
\end{align*}
Requests arrive in chronological order. To satisfy the constraint $C_{platform}$ (e.g. the maximum proportion of displaying ads during a day of the platform ), if the system exposes too many ads during early period, it should expose fewer ads later. Hence the above problem is naturally a sequential decision-making problem.


\subsubsection{Problem Formulation}
To solve such a sequential decision-making problem, one typical method is to model it as MDP \cite{cai2017real} or CMDP \cite{wu2018budget}. and then use reinforcement learning techniques to solve the problem. In practice, we cannot acquire and make accurate predictions of the environmental information like $v(b(q_i, d_j;\theta), q_i, d_j)$ and $w(b(q_i, d_j;\theta), q_i, d_j)$ aforehand, thus we resort to model-free reinforcement learning techniques.
However, since there exist both \emph{platform-level} and \emph{request-level} constraints, the traditional CMDP \cite{altman1999constrained} cannot be directly applied here. We propose a special CMDP which we term as CMDP with per-state constraint (psCMDP). Formally a psCMDP can be defined as a tuple $(S, A, R, P, C_{T}, C_{S})$. Comparing to the original CMDP \cite{altman1999constrained}, we see that the difference here is that for each trajectory $\tau$, psCMDP needs to satisfy not only the trajectory-level constraint $C_{T}$: 
\begin{equation}
J_{C_T}(\pi) = \sum_{t=0}^{T}\gamma^{t} C_T(s_t, \pi(s_t), s_{t+1}) \le u_T
\tag{9} 
\end{equation}
but also the per-state constraint $C_{S}$ over each request:
\begin{equation}
J_{C_S}(\pi) = C_S(s_t, \pi(s_t), s_{t+1}) \le u_S, \forall (s_t, a_t, s_{t+1}) \in \tau
\tag{10} 
\end{equation}
where $C_{S}: S \times A \times S \rightarrow \mathbb{R}$ and $u_S$ is the upper bound of $C_S$. So the set of feasible stationary policies for a psCMDP is:
\begin{equation}
\Pi_{ps} = \{\pi \in \Pi : J_{C_T}(\pi) \le u_T \} \cap \{\pi \in \Pi : J_{C_S}(\pi) \le u_S \}
\label{cmdp_p1}
\tag{11} 
\end{equation}

The components of a psCMPD are described in details as follows:
\begin{itemize}
    \item $S$: The state should reflect both the environment and the constraints in principle. In our settings, we consider the following statistics for $s_t$: 1) information related to the the current request  $q_i$, e.g., features of the candidate ads; 2) system context information, e.g., the number of ads exposed up to time $t$.
    \item $A$: Considering the system picks out products for each request by the score of all the products, 
    we adjust all the ad candidates' score of a request at once. Accordingly, we denote 
    $a_t = ({\eta_1^{q_i}, \eta_2^{q_i}, ..., \eta^{q_i}_{N^{d}}})$, where $\eta_j^{q_i}$ is the coefficient of the $j$-th ad $d_j$ for request $q_i$, where $1 \le j \le N^{d}$.
    \item $R$: $R(s_t, a_t) = \sum_{d \in \mathcal{D}_{s_t}^{a_t}} v(s_t, d)$, where $a_t$ is the score adjustment action in state $s_t$, $D_{s_t}^{a_t}$ is the set of ads finally exposed in $s_t$ and $v(s_t, d)$ is the revenue value of displaying ad $d$ in $s_t$, We set $v(s_t, d)$ as the Generalized Second-Price after the actual sorting of the advertising items and recommended items.
    \item $P$: State transition models the dynamics of requests visiting sequence and system information changes. The effect of $a_t$ on state transitions is reflected in: Different $a_t$ would lead to different ads in $s_t$, which would also affect the total number of ads that have been shown (which is a component of $s_{t+1}$). Similar ways of modeling state transitions have also been adopted previously in \citet{cai2017real}, \citet{jin2018real} and \citet{wu2018budget}.
\end{itemize}
Specifically, for the constraints:
\begin{itemize}
\item $C_T$: It is defined as the platform level constraint $C_{platform}$ - the advertising exposure constraint over a \textbf{day (trajectory)}, and set discount factor $\gamma$ to 1, $C_T: \frac{\Sigma_{1\le i \le M} N_{i}^{d}}{\Sigma_{1\le i \le M} N_{i}} \le \alpha$. 
\item $C_S$: It is defined as the request level constraint $C_{request}$ - the advertising exposure constraint over each \textbf{request (state)}, and set discount factor $\gamma$ to 1, $C_S: \frac{N_{i}^{d}}{N_{i}} \le \beta, \forall i \in \mathbb{N}$.
\end{itemize}
With all the definitions above, an optimal policy $\pi^*$ is defined as follows:
\vspace{-0.4cm}
\begin{equation} 
\begin{aligned}
\pi^* &= \underset{\pi \in \Pi_{ps}}{\arg\max}\ E[ \sum_{t = 0 }^{M} R(s_t, \pi(s_t))]\\
&= \underset{\pi \in \Pi_{ps}}{\arg\max}\ E[ \sum_{t = 0 }^{M}\sum_{d \in \mathcal{D}_{s_t}^{a_t}|a_t = \pi(s_t)}v(s_t, d)]
\end{aligned}
\tag{12}
\end{equation}

Our problem also shares similarity with contextual bandit problem well-studied in the E-commerce literature. However, one major difference is that contextual bandit mainly studies choosing an action from the fixed and known set of possible actions, such as deciding whether to display advertisements, and which locations to display advertisements \cite{badanidiyuru2014resourceful, wu2015algorithms, agrawal2016efficient, tang2013automatic}. In our case, however we have to adjust the scores (which are continuous variables) of hundreds of millions of items to maximize the total reward, which drives us to model the state transitions explicitly. The other reasons for adopting RL instead of contextual bandit are as follows: 1) \citet{wu2018budget} show that modeling trajectory constraints into RL will lead to higher profits since RL can naturally track the changes of constraints in a long run and make longer term decisions. 2), \citet{hu2018reinforcement} further confirm that RL methods can bring higher long-term returns than contextual bandit methods for ranking recommended products in e-commerce.

\subsection{Solution: Constrained Two-level Reinforcement Learning}
We propose a constrained two-level reinforcement learning framework to address the constrained advertising optimization problem. The overall structure is shown in Fig. \ref{temporal} and the framework is described in  Algorithm 1. We split the entire trajectory into a number of sub-trajectories. The optimization task of the higher level is to learn the optimal policy $\pi_{C_T}$ for selecting constraints for different sub-trajectories to maximize long-term revenue while ensuring that the constraint over the whole trajectory $C_T$ is satisfied (Algorithm 1, Line \ref{algo1_4}). Given a sub-trajectory constraint $C_{ST}$ from the higher level, the lower level is responsible for learning the optimal policy $\pi_b(s_t;C_{ST})$ over its sub-trajectory while ensuring that both the sub-trajectory constraint $C_{ST} $ and the per-state constraints $C_S$ are satisfied (Algorithm 1, Line \ref{algo1_2} - \ref{algo1_3}). In this way, the original psCMDP optimization problem is simplified by decoupling it into the two independent optimization sub-problems. 

By decoupling the adaptive exposure learning problem in such a two-level manner, we have higher adaptability and can quickly make response to dynamically changing e-commerce environments. 
This property is critical in online e-commerce environments since slower response would result in significant amount of monetary loss to the company. First, the platform-level constraint may vary frequently due to the company's business strategic change. In this case, the lower level policies we have learned can be reused and only the higher level policy needs to be retrained. Second, the recommendation system or other components of the adverting system may change frequently, and our adjustment policy needs to be updated accordingly. In this case, we only need to retrain the lower level policies while the higher level part can be retained.

\begin{algorithm}[tp]
\caption{Constrained Two-level Reinforcement Learning}
\begin{algorithmic}[1]
\STATE Given: constraints $C_T$ and $C_S$, an off-policy RL algorithm $Algo$ \label{algo1_1}
\STATE Initialize the state-level constraint set $\mathcal{C}_{S}$ based on $C_S$ and the environmental information. \label{algo1_2}
\STATE Lower Level: Under the premise of satisfying the constraint $C_S$, train the state-level behavior policy set according to $\mathcal{C}_{S}$ and use Constrained Hindsight Experience Replay (introduced in section 3.3.1) to speed up the training process. \label{algo1_3} 
\STATE Higher Level:  According to the state-level behavior policy set, assign constraints on different sub-trajectory to maximize the expected long-term advertising revenue while satisfying the trajectory-level constraint $C_T$.
\label{algo1_4}
\end{algorithmic}
\end{algorithm}

\subsubsection{Lower Level Control}\label{low_level}
In the lower level, we have to address the sub-problem of learning an optimal advertising policy under a particular sub-trajectory constraint $C_{ST}$ provided by the higher level part.
In our approach, we convert the constraint of each $C_{ST}$ straightforward into state level constraints to do more precise manipulating. We simplify the sub-trajectory optimization problem at the cost of sacrificing the policy optimality by treating the sub-trajectory constraint $C_{ST}$ as a state-level constraint. It is obvious that the original state-level constraint $C_{S}$ is more strict than $C_{ST}$. Thus, we can easily combine $C_{ST}$ and $C_{S}$ into a single state-level constraint $C_{ST}$.
Obviously once the state level constraint is satisfied, the higher level constraint would be satisfied at the same time. Thus, given a sub-trajectory constraint $C_{ST}$ by the higher level policy, we can directly optimize the lower level policy at the state level. One natural approach in CMDP is to guide the agent's policy update by adding an auxiliary value related to the per-state constraint to each immediate reward (Equation. \ref{add_reward}). And, during policy update, both the current and the future penalty values are considered. However, in our lower level, since each transition satisfies the constraint $C_S$ independently, each action selection does not need to consider its future per-state constraints $C_S$.

\subsubsection*{Enforce Constraints with Auxiliary Tasks} Considering the above reasons, we propose a method similar to auxiliary tasks \cite{jaderberg2016reinforcement} by adding an auxiliary loss function based on per-state constraints. We use $L_{RL}$ and $L_{C_{S}}$ to denote the RL loss function and the per-state constraint $C_S$ loss function respectively.
During  training, the policy is updated towards the direction of minimizing the weighted sum of the above two: 
\begin{equation}
\min\ L'(\theta) = w_1 \times L_{RL}(\theta) + w_2 \times L_{C_{S}}(\theta)
\tag{13} 
\end{equation}
where $w_1$ and $w_2$ are the weights, $\theta$ are the parameters of the value network.
For example, for critic part in DDPG \cite{lillicrap2015continuous}, the original critic loss function is:
\begin{equation}
L_{RL}(\theta) = [r + \gamma Q(s_{t+1}, \pi(s_{t+1}|\theta^\pi);\theta') - Q(s_t, a_t;\theta)]^2
\tag{14}  
\end{equation} 
and the additional loss function for per-state constraints can be defined as follows:
\begin{equation}
L_{C_S}(\theta) = [Q(s_t, a_t;\theta') + \delta(s_t, a_t, s_{t+1} | C_S) - Q(s_t, a_t;\theta)]^2
\tag{15}  \end{equation}
where $\theta$, $\theta^\pi$ and $\theta'$ are the online critic network parameters, actor network parameters and the target critic network parameters respectively and $\delta(s_t, a_t, s_{t+1}|C_S)$ is the function of $C_S$. The value of $\delta(s_{t}, a_{t}, s_{t+1}|C_S)$ is used to control the degree that Q-function is updated when the constraint is violated. 
For example, when we want to limit the pvr of each request close to 0.4, we can set $\delta(s_{t}, a_{t}, s_{t+1}|C_s)= - (pvr_t - 0.4)^2$, where $pvr_t$ is the pvr value of request $q_t$. 
Intuitively, the more $(s_t, a_t, s_{t+1})$ deviates from the target pvr, the more its corresponding Q-value will be decreased. 
Similar techniques have also been used to ensure the optimality of the expert demonstrations by using a margin classification loss \cite{hester2018deep}.

\begin{figure*}[!ht]
\centering
\begin{minipage}{0.33\linewidth}
\includegraphics[width=1\textwidth]{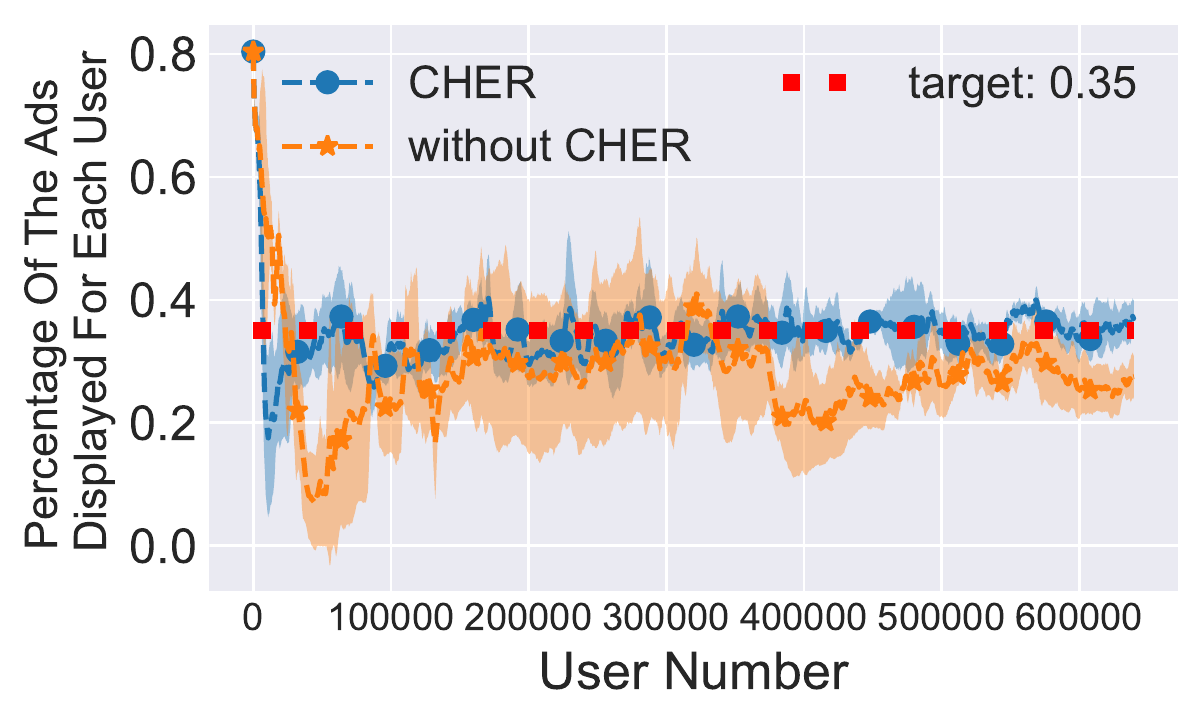}
\captionof*{figure}{(a): The objective of the policy is:  0.35}
\end{minipage}
\begin{minipage}{0.33\linewidth}
\includegraphics[width=1\textwidth]{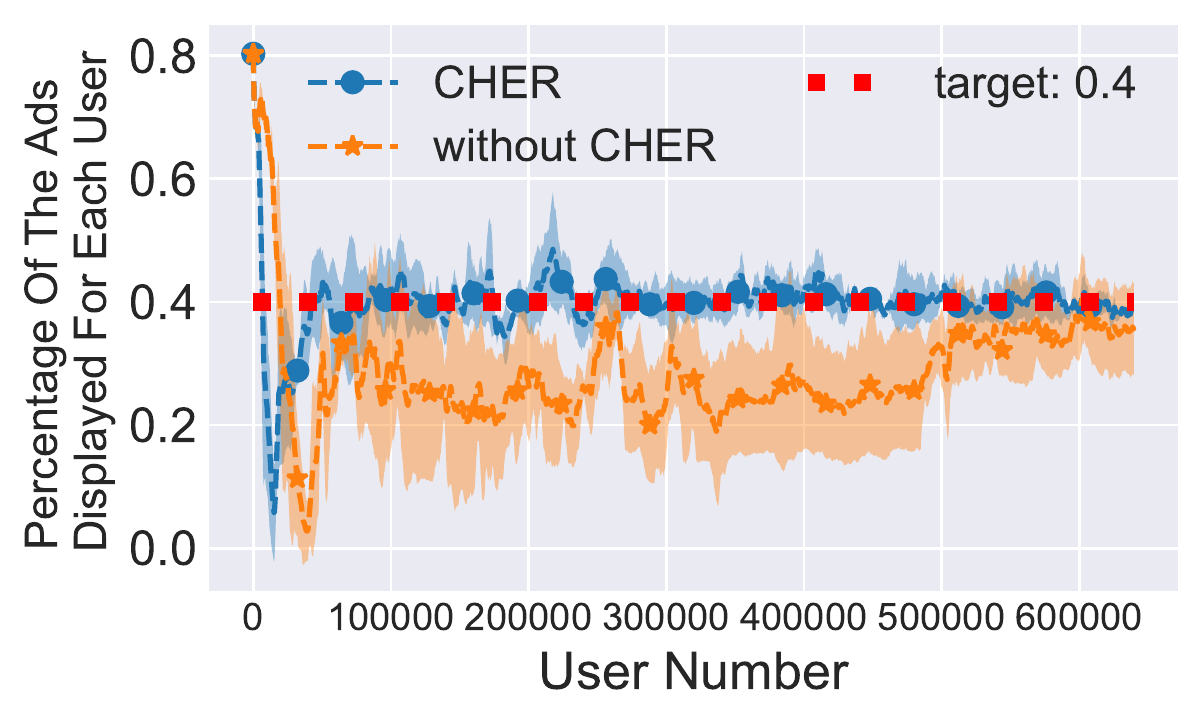}
\captionof*{figure}{(b): The objective of the policy is:  0.4}
\end{minipage}
\begin{minipage}{0.33\linewidth}
\includegraphics[width=1\textwidth]{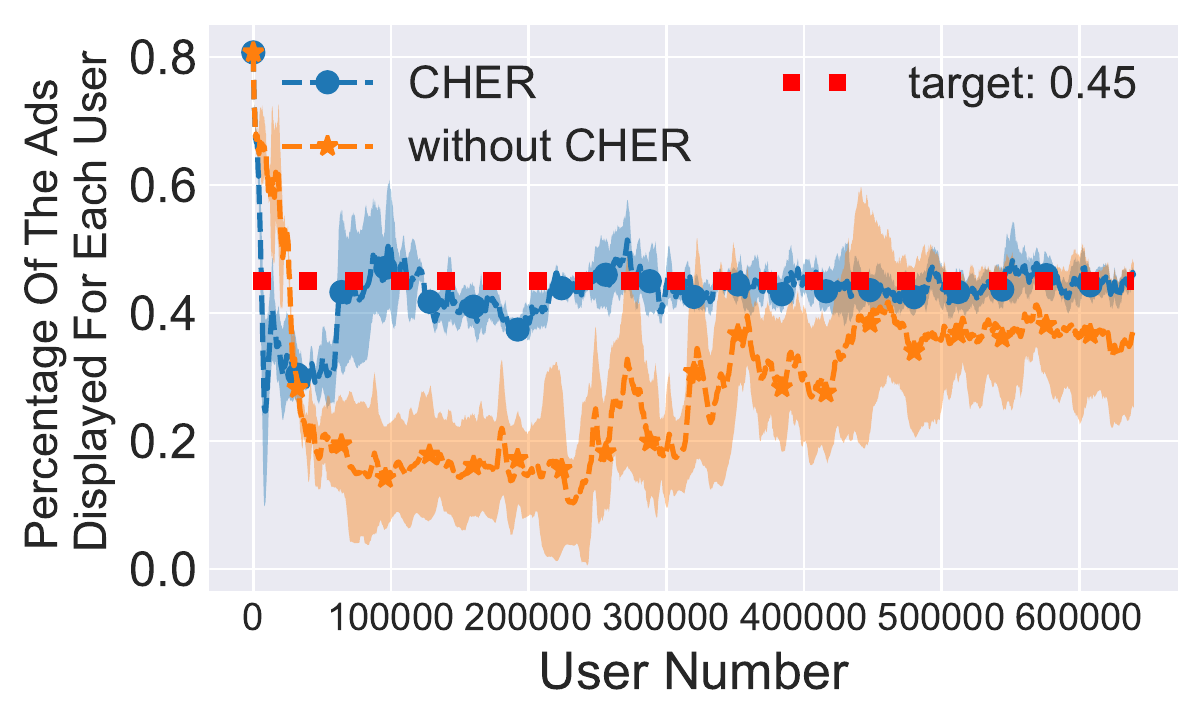}
\captionof*{figure}{(d): The objective of the policy is:  0.45}
\end{minipage}
\caption{Learning Curves: DDPG with CHER Compared with DDPG without CHER.}
\label{fig_cher}
\end{figure*}




\begin{algorithm}[tp]
\caption{Constrained Hindsight Experience Replay}
\begin{algorithmic}[1]
\STATE Given: a state-level constraints set $\mathcal{C}_{S}$, epoch number $M^e$, training repeat number $N$.
\STATE Initialize: replay buffer $B$.
\FOR {$epoch = 1, M^e$} 
\STATE Sample a constraint $c_s$ in $\mathcal{C}_{S}$ and initial state $s_0$.
  \WHILE{$s_t$ is not terminating state} \label{algo2_1}
    \STATE Sample an action $a_t$ using the behavioral policy: $a_t \leftarrow \pi_b(s_t; c_s)$. \label{algo2_2}
    \STATE Execute $a_t$ and observe  state $s_{t+1}$, reward $r_{t}$.\label{algo2_3}
    \STATE Store the transition $(s_t, a_t, r_t, s_{t+1}, c_s)$ in $B$.\label{algo2_4}
  \ENDWHILE
  
  \FOR {$t = 1, N$}
    \STATE Sample a transition $(s_t, a_t, r_t, s_{t+1}, c_s)$ from $B$, and train $\pi_b$ using add constrained: $L' = L_{RL} + L_{c_{s}}$.
        \FOR {$ c_{s'} \in \mathcal{C}_{S}$} 
          \STATE use  transition $(s_t, a_t, r_t, s_{t+1}, c_{s'})$ to train $\pi_b(s_t; c_{s‘})$ by $L' = L_{RL} + L_{c_{s'}}$. \label{algo2_5}
        \ENDFOR
  \ENDFOR
\ENDFOR
\end{algorithmic}
\label{lower_lev}
\end{algorithm}

\subsubsection*{Constrained Hindsight Experience Replay} To increase the sample efficiency, we propose leveraging the idea of hindsight experience replay (HER) \cite{andrychowicz2017hindsight} to accelerate the training of optimal policies for different sub-trajectory constraints. HER relieves the problem of sample inefficiency in DRL training by reusing transitions, which can be obtained by using different goals to modify reward. We extend this idea to propose the constrained hindsight experience replay (CHER). Different from HER, CHER does not directly revise the reward. Instead, it uses different constraints to define the extra loss $L_{C_S}$ during training. The overall algorithm for training lower level policies under CHER is given in Algorithm \ref{lower_lev}. When we learn a policy to satisfy constraint $c_{s}$ (specific constraints on each state), it obtains the transition: $(s_t, a_t, r_t, s_{t+1}, c_{s})$ (Algorithm \ref{lower_lev}, Line\ref{algo2_1} - \ref{algo2_4}). We can replace $c_{s}$ with another constraint $c_{s'}$ and then reuse those samples $(s_t, a_t, r_t, s_{t+1}, c_{s'})$ and $\delta(s_t, a_t, s_{t+1}|c_{s'})$ to train a policy satisfying constraint $c_{s'}$ (Algorithm \ref{lower_lev}, Line\ref{algo2_5}).

\subsubsection{Higher Level Control}

The higher level task is to determine trajectory-level constraints for each sub-trajectory to maximize the expected long-term advertising revenue while satisfying the original trajectory constraint $C_T$. \footnote{By satisfying the state-level constraint in the lower level, we reduce the optimization problem of the higher level into an optimization problem that only needs to consider the trajectory-level constraint.} At each decision point, the higher level policy (we term as constraint choice policy, CCP) selects a constraint for the next sub-trajectory, and the corresponding lower level policy takes over and determines the ads adjustment score for each request within that sub-trajectory. After the lower level policy execution is finished, the accumulated revenue over that sub-trajectory is returned to the higher level policy as its abstraction immediate reward $r^{ab}$. The above steps repeat until we reach the end of the trajectory, and then we obtain the actual percentage of of ads displayed over the whole trajectory, which can be compared with the trajectory constraint as an additional reward $r^\tau$ with weighted $w_2$:
\vspace{-0.1cm}
\begin{equation}
r^{ab'}= r^{ab} + w_2 \times r^\tau
\tag{16}  
\end{equation}
Similar to WeiMDP \cite{geibel2006reinforcement}, we use DQN \cite{mnih2015human} for the higher level policy optimization. More advanced RL techniques (such as CPO \cite{achiam2017constrained}, SPSA \cite{prashanth2016variance}) can be applied as well. Note that our higher level control is similar to the temporal abstraction of hierarchical reinforcement learning (HRL) \cite{bacon2017option,kulkarni2016hierarchical}.  However, in contrast to learning how to switch option \cite{bacon2017option} and alleviate the sparse reward problem\cite{kulkarni2016hierarchical} in HRL,  our work leverage the idea of hierarchy to decompose different constraints into different levels.



\section{Experiments}

\subsection{Experimental Setup}
Our experiments are conducted on a real dataset of the Chinese largest E-commerce platform Taobao, and the data collection scenario is consistent with the problem description in Section 3. 
In Fig. \ref{fig_ill}, we have demonstrated that the score adjustment of a product produced by the advertising system does not affect the selection and scoring of the candidate products produced by the recommendation system. It only influences the relative ranking of the ads compared with the recommendation products and affects the final mixed sorting results. Therefore, the online data collected from platform can be reused to evaluate the effects of score adjusting through resorting the mixture of the original recommended products and the regraded ad products. Similar settings can be found in related work \cite{cai2017real,jin2018real,perlich2012bid,zhang2014optimal}. 
Specificailly, we replay the users' access logs in chronological order to simulate the users' requests. The state of our psCMDP is represented by integrating the features of all candidate ads and the system contextual information. The action is defined as the score adjusting ratios for candidate ads. Finally, the reward and the satisfaction condition of each constraint are calculated accordingly following the definition in Section 3.2. 
All details can be found in the appendix.

\subsection{Does CHER improve performance?}
\begin{figure}
\begin{minipage}[b]{0.4\linewidth}
\includegraphics[width=1\textwidth]{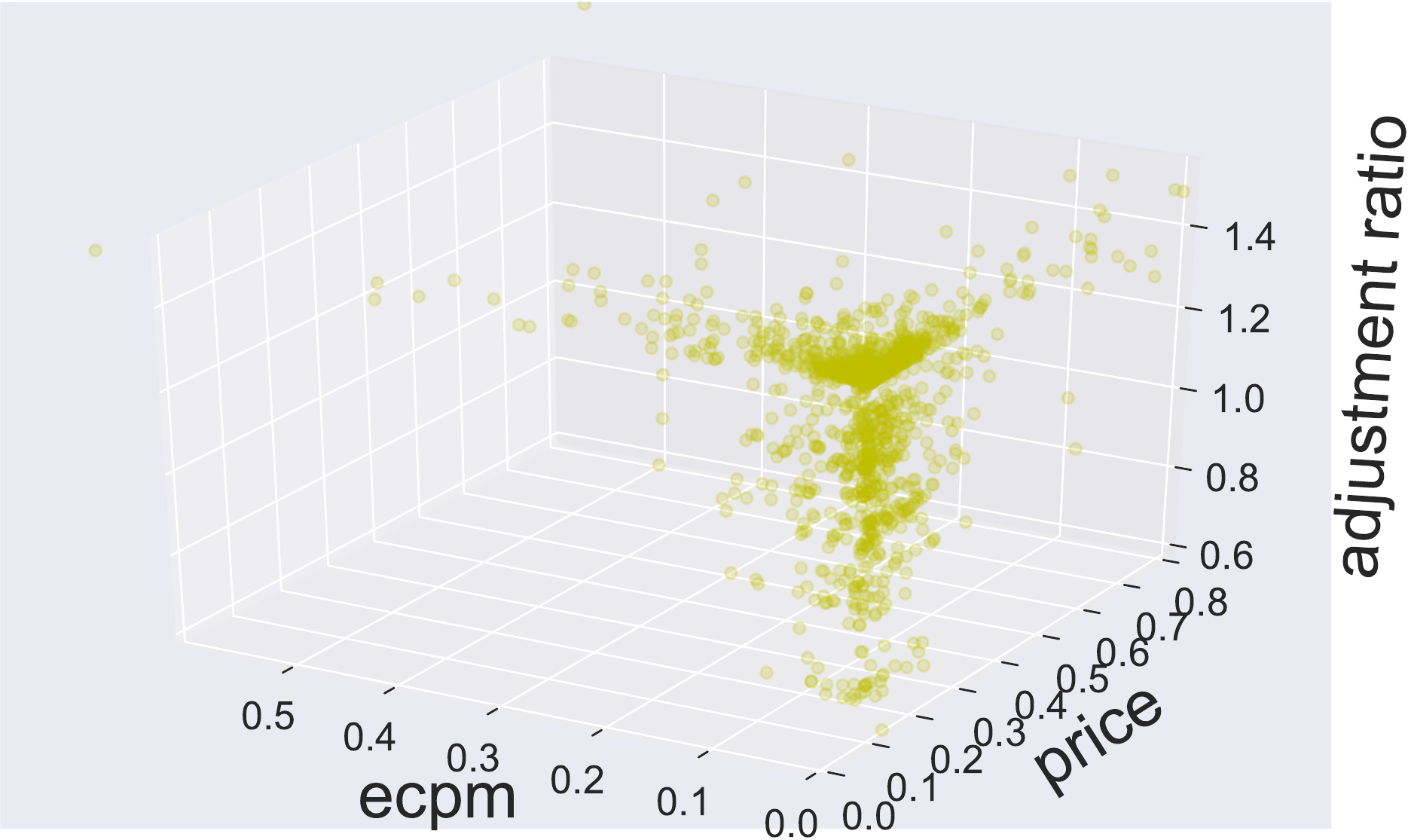}
\captionof*{figure}{(a)}
\end{minipage}
\hspace{5pt}
\begin{minipage}[b]{0.37\linewidth}
\includegraphics[width=1\textwidth]{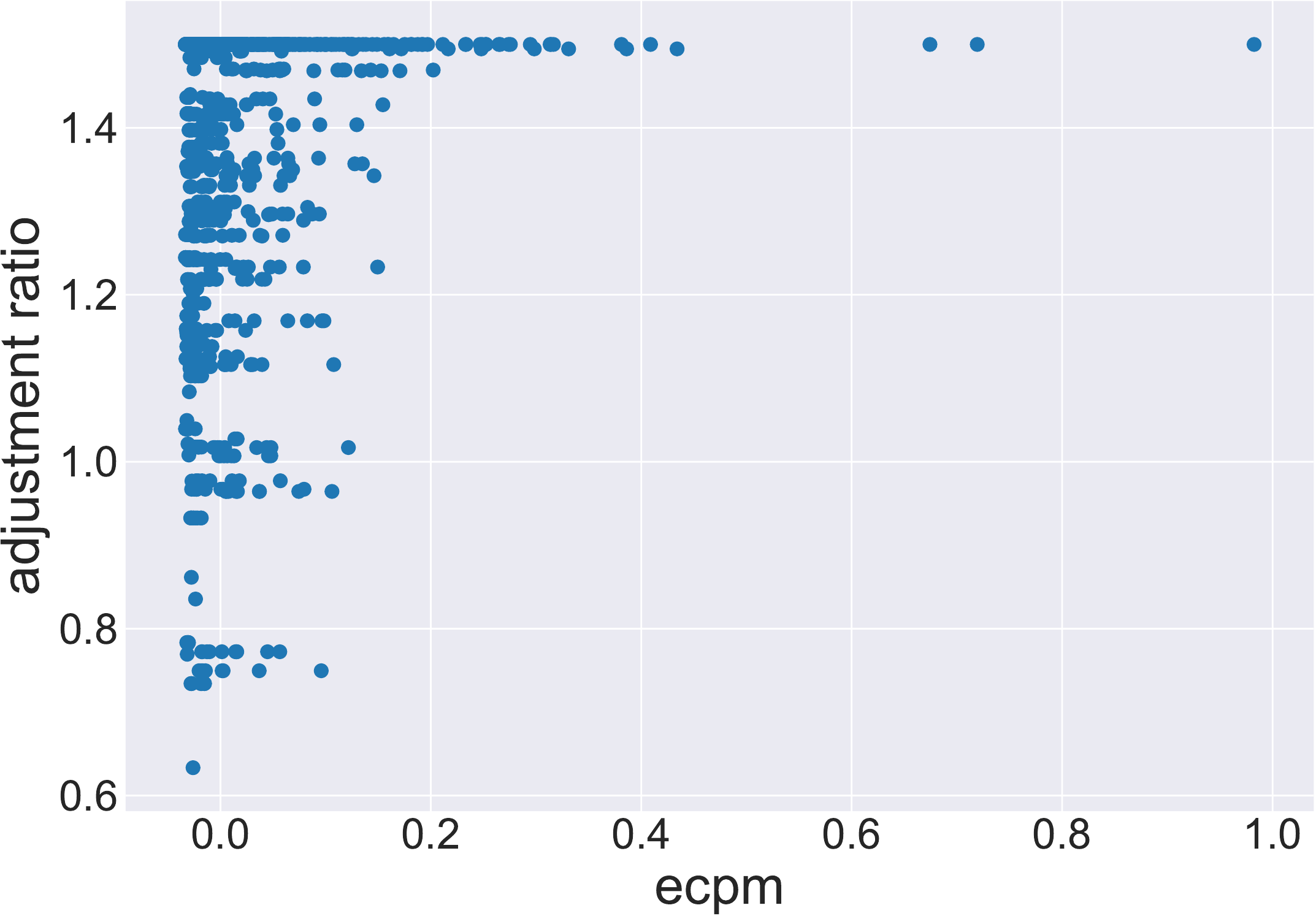}
\captionof*{figure}{(b)}
\end{minipage}
\vspace{-15pt}
\caption{The relationship between ratio and the value of advertising products. (a): The relationship between ratio and commodity ecpm, price. (b): The relationship between ratio and commodity ecpm}
\label{lower_level_action}
\end{figure}

To verify the effectiveness of using CHER, we compare the impact of using CHER on the learning speed and stability with a baseline DDPG \cite{lillicrap2015continuous} under the same network structure and parameters.$^4$ Suppose $C_S$ is the number of exposure ads for each request, and cannot exceed 5, so we set $\mathcal{C}_{S}$ to be consisting of 5 constraints. Each goal in $PV\hspace{-1pt}R_i = 0.3, 0.35 , 0.4, 0.45, 0.5$ represents the expected average number of ads exposed per request. Intuitively, we can use the constraint as part of the input and use a network to satisfy all constraints \cite{andrychowicz2017hindsight}. However, considering the learning stability, we use 5 different networks to satisfy different constraints. Since we use DDPG, we add $L_{C}$ to $L_{Critic}$ during Critic training, 
\begin{equation}
L_{Critic}' = L_{Critic} + w \times L_{C}
\tag{17}  \end{equation}
\begin{equation}
L_{C} =  E_i[(-|PV\hspace{-1pt}R_i - {PV\hspace{-1pt}R}^{t}| + Q(s, a;\theta') - Q(s, a;\theta))^2]
\tag{18}  \end{equation}
where $w$ is set to 10, and $PV\hspace{-1pt}R_i$ is the percentage of ads exposed for $i$-th request $q_i$. 
We set up 4 different random seeds, and the experiment results are shown in Fig. \ref{fig_cher}. We only show the results of $PV\hspace{-1pt}R_i = 0.35 , 0.4, 0.45$ due to the space limit. The criterion for an algorithm will be better if its result is closer to the target constraint. We can find that, under different constraints, DDPG with CHER is better than DDPG in terms of training speed, achieving and stabilizing around constraints. In order to understand the rationality of the policy after training, we randomly sampled some user visits in the dataset. By recording the actions of adjusting scores of the advertisements in these user visits (Fig. \ref{lower_level_action}), our approach learning can be intuitively understood as follows: if the advertising product has higher value (ecpm, price), then its score is adjusted higher.

\begin{figure}[!pt]
\centering
\begin{minipage}[b]{0.5\linewidth}
\includegraphics[width=1\textwidth]{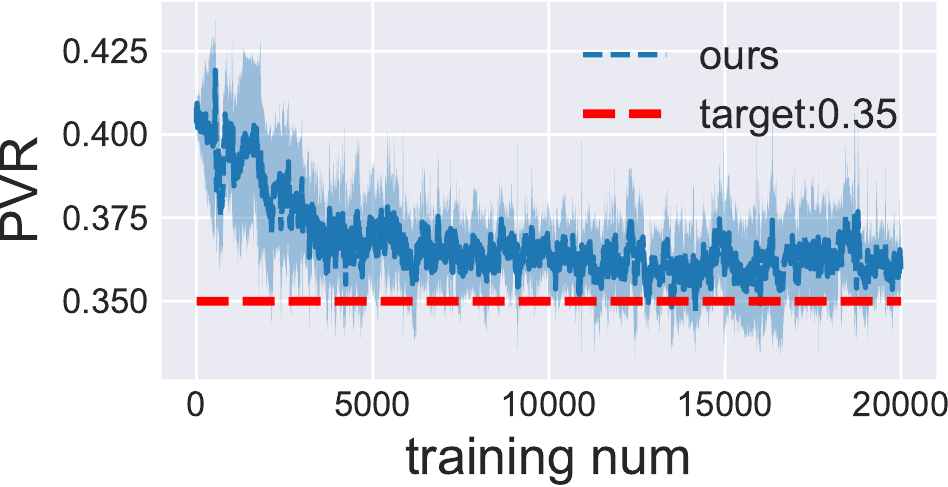}
\captionof*{figure}{(a): PV\hspace{-1pt}R.}
\end{minipage}
\begin{minipage}[b]{0.49\linewidth}
\includegraphics[width=1\textwidth]{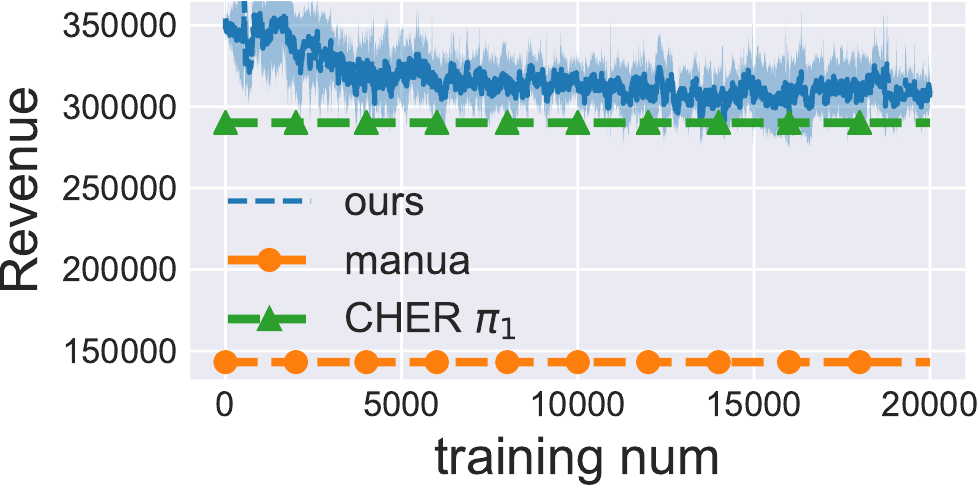}
\captionof*{figure}{(b): Revenue.}
\end{minipage}
\vspace{-22pt}
\caption{Learning Curves Compared with Policy $\pi_1$.}
\label{crl_0_35}
\vspace{8pt}

\begin{minipage}[b]{0.5\linewidth}
\includegraphics[width=1\textwidth]{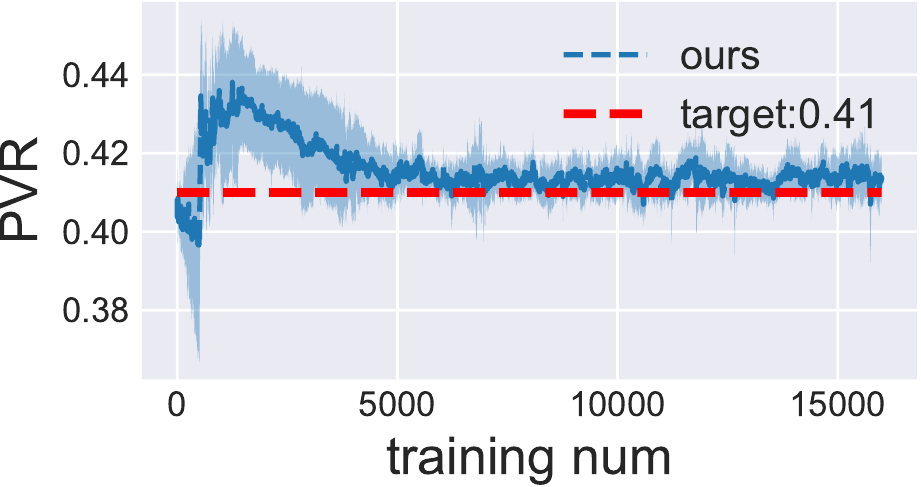}
\captionof*{figure}{(a): PV\hspace{-1pt}R.}
\end{minipage}
\begin{minipage}[b]{0.49\linewidth}
\includegraphics[width=1\textwidth]{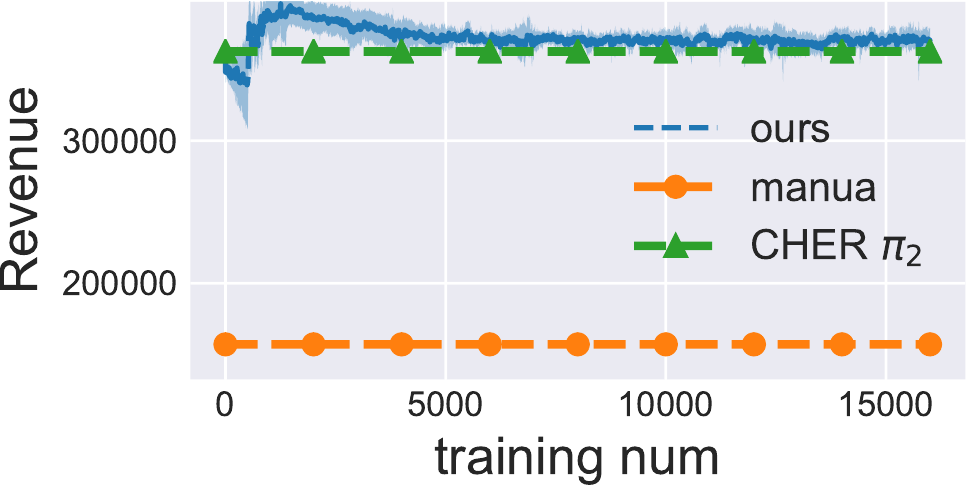}
\captionof*{figure}{(b): Revenue.}
\end{minipage}
\vspace{-22pt}
\caption{Learning Curves Compared with Policy $\pi_2$.}
\label{crl_0_41}
\vspace{8pt}

\begin{minipage}[b]{0.5\linewidth}
\includegraphics[width=1\textwidth]{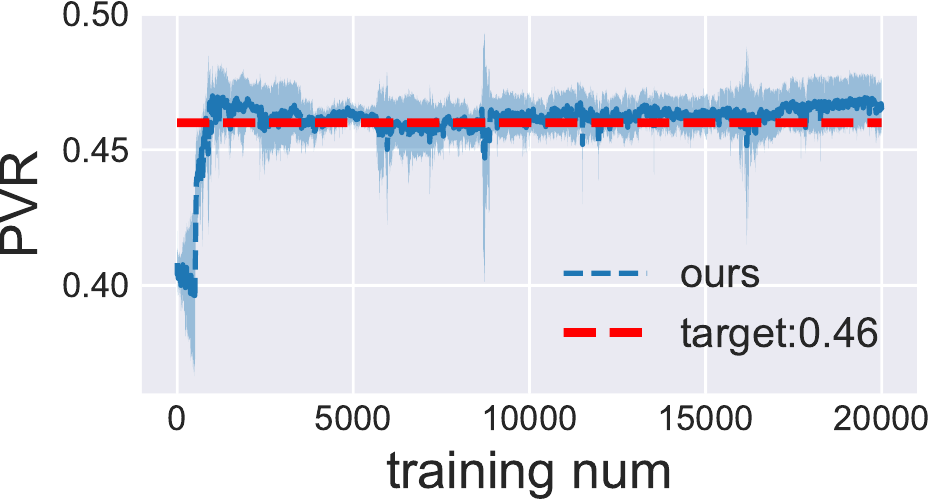}
\captionof*{figure}{(a): PV\hspace{-1pt}R.}
\end{minipage}
\begin{minipage}[b]{0.48\linewidth}
\includegraphics[width=1\textwidth]{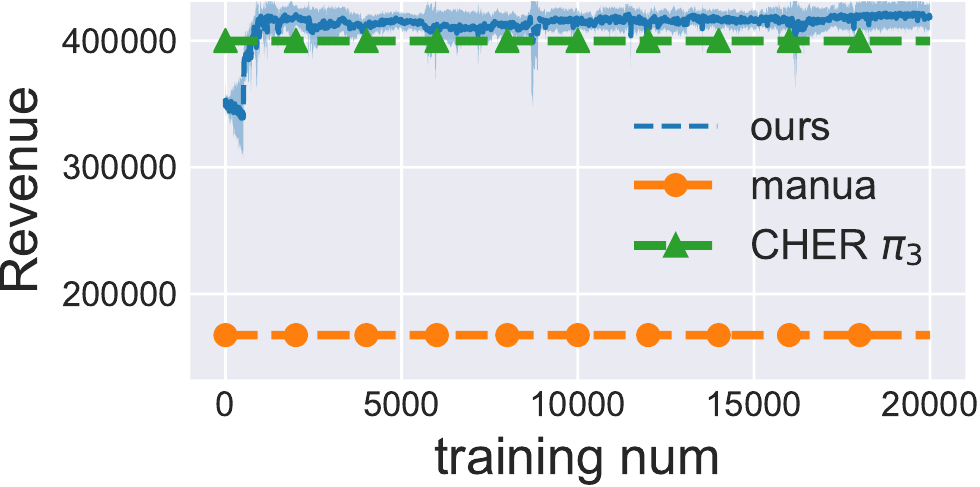}
\captionof*{figure}{(b): Revenue.}
\end{minipage}
\vspace{-10pt}
\caption{Learning Curves Compared with Policy $\pi_3$.}
\label{crl_0_46}
\vspace{-15pt}
\end{figure}


\begin{table*}[!pt]
\centering
\fontsize{8}{10}\selectfont
\caption{The Performance of Our Approach in One Day.} %
\begin{tabular}{| c | c c | c | c c | c | c c |}\hline
Policy &\multicolumn{2}{c|}{performance}&Policy &\multicolumn{2}{c|}{performance}&Policy &\multicolumn{2}{c|}{performance}\\
&PVR&Revenue&&PVR&Revenue&&PVR&Revenue\\
\hline
target & 0.35& -  & target & 0.41 & - & target & 0.46 & -\\
\hline
manual & 0.3561 & 143121 (\textbf{100\%}) & manual & 0.4179 & 157120 (\textbf{100\%}) & manual & 0.4640 & 167489 (\textbf{100\%})\\
\hline
CHER $\pi_1$ & 0.3558&290260 (\textbf{202.8\%}) & CHER $\pi_2$ & 0.4100 & 362676 (\textbf{230.8\%}) & CHER $\pi_3$ & 0.4608&399712 (\textbf{238.6\%})\\
\hline
CCP & 0.3576 &308108 (\textbf{215.3\%}) & CCP & 0.4141 & 370914 (\textbf{236.1\%}) & CCP & 0.4673 & 420119 (\textbf{250.8\%}) \\

\hline
\end{tabular}
\label{tabe2_per}
\end{table*}


\subsection{Verify the Effectiveness of Constrained Two-level Reinforcement Learning}
In order to verify the two-level structure can bring about an increase in revenue, we compare the performance of different methods under different \emph{platform-level} constraints PVR=0.35, 0.41, 0.46 (the upper bound of the advertising rate of each day is 0.35, 0.41, 0.46 respectively,  $C_T$=0.35, 0.41, 0.46) with the state level constraint fixed ($PVR_i=0.5$, the upper bound of the advertising rate of each request is 0.5, $C_S=0.5$).
Since we are considering a new adaptive exposure mechanism, there are no existing approaches suitable for comparison. In our paper, we consider the following two approaches as baselines: 1). manual: the score of an advertisement is manually adjusted by human experts. 2). CHER + DDPG: a previous trained model of Section 4.2. It corresponds to a policy of using a fixed \emph{request-level} constraint for the whole trajectory without adaptive adjustment. Since the performance of DDPG varies a lot 
, we add a complementary CHER to DDPG and use this optimized approach (CHER+DDPG) to attain a more stable $PVR_i$.


\begin{table*}[!pt]
\centering
\fontsize{8}{10}\selectfont
\caption{The Performance of Our Approach for each Hour.} %
\begin{tabular}{|c|c c c|c c c|c c c|}
\hline
\multirow{2}*{Hour} & \multicolumn{3}{c|}{Revenue} &\multicolumn{3}{c|}{PVR} &\multicolumn{3}{c|}{Revenue / PVR} \\
& DDPG+CHER $\pi_1$ &  CCP & CCP - $\pi_1$ $*$ & DDPG+CHER $\pi_1$ & CCP & CCP - $\pi_1$ $*$ & DDPG+CHER $\pi_1$ & CCP & CCP - $\pi_1$ $*$\\
\hline
8& 11556  &  15845 & 4289  & 0.01280837  & 0.01590289 & 0.003094520   &902222  & 996359  &94137\\
9& 15595  &  23422 & 7827  & 0.0162089   & 0.02192751 & 0.005718610   &962125  & 1068155 &106030\\
10& 20157 &  28979 & 8822  & 0.0184266   & 0.02321300 & 0.004786400   &1093907 & 1248395 &154487\\
11& 18221 &  24739 & 6518  & 0.01880709  & 0.02246692 & 0.003659830   &968836  & 1101130 &132293\\
12 &16777 &  18646 & 1869           & 0.01794808  & 0.01895375 & 0.001005670   &934751  & 983763  &49011\\
-& -  &  - & -  & -  & - & -   &-  & -  &-\\
15 &18129 &  16023 & -2106          & 0.02096899  & 0.01851524 & -0.00245375   &864562  & 865395  &832\\
16 &22913 &  20450 & -2463          & 0.02233828  & 0.01964052 & -0.00269776   &1025727 & 1041214 &15486\\
-& -  &  - & -  & -  & - & -   &-  & -  &-\\
17 &12919 &  11432 & -1487          & 0.01914268  & 0.01718366 & -0.00195901   &674879  & 665283  &-9596\\
18 &11424 &  9786  & -1638          & 0.01633943  & 0.01428198 & -0.00205745   &699167  & 685199  &-13968\\
19 &11586 &  10081 & -1505          & 0.01570854  & 0.01391865 & -0.00178989   &737560  & 724280  &-13280\\
-& -  &  - & -  & -  & - & -   &-  & -  &-\\
22 &18362 &  15391 & -2971          & 0.02780465  & 0.02398777 & -0.00381688   &660393  & 641618  &-18774\\
23 &12720 &  10584 & -2136          & 0.02291417  & 0.01988751 & -0.00302666   &555115  & 532193  &-22921\\
\hline
\end{tabular}
\label{tabe3_per}
\footnotesize{\\ notice that CCP - $\pi_1$ $*$ is the performance difference of CCP and $\pi_1$ under different evaluation indicators (e.g. Revenue, PVR, Revenue/PVR)}
\end{table*}


\subsubsection*{Does higher level control improve performance?}To distinguish different policies in the behaviour policy set, we use $\pi_0$, $\pi_1$, $\pi_2$, $\pi_3$, $\pi_4$ to refer to the different lower level policies (DDPG+CHER) previous trained in Section 4.2 under different \emph{platform-level} constraints ${PV\hspace{-1pt}R} = 0.3, 0.35, 0.4, 0.45, 0.5$. 
The temporal abstraction value is set to 1 hour, which means the higher level CCP makes a decision per hour.\footnote{In fact, a more fine-grained decomposition can lead to a better performance. However, we simply set the minimum temporal unit to 1 hour here to make the analysis of the improvement of CCP easier.}
After selecting the sub-trajectory constraint, the behavior policy of the state level is activated for adjusting ads' scores in the flowing hour with the sub-trajectory constraint fixed. 
In our experiments, We combine the double DQN architecture with the dueling structure$^4$ to train CCP. 
The state of CCP consists of hourly information, such as the timestamp, hourly eCPM, PV\hspace{-1pt}R from 00:00 to current time.
The objectives of the higher level policy are: (1) achieving approximately the same number of exposure ads with target $PV\hspace{-1pt}R$; (2) improving revenue as much as possible. 
Detailed results are shown in Table. \ref{tabe2_per} and Fig. \ref{crl_0_35} - \ref{crl_0_46}, in which we see the CCP can increase the revenue of each day compared to the manual and DDPG+CHER policies under the same constraint $C_T$. 
Therefore, we demonstrate that our approach learns to expose different numbers of ads in different time periods, which means that more ads are exposed when the value of the ads in a request is higher and fewer ads are shown in other time slots.

\begin{figure}[!pt]
\setlength{\belowcaptionskip}{-0.43cm}
\centering
\begin{minipage}[b]{0.9\linewidth}
 \includegraphics[width=1\textwidth]{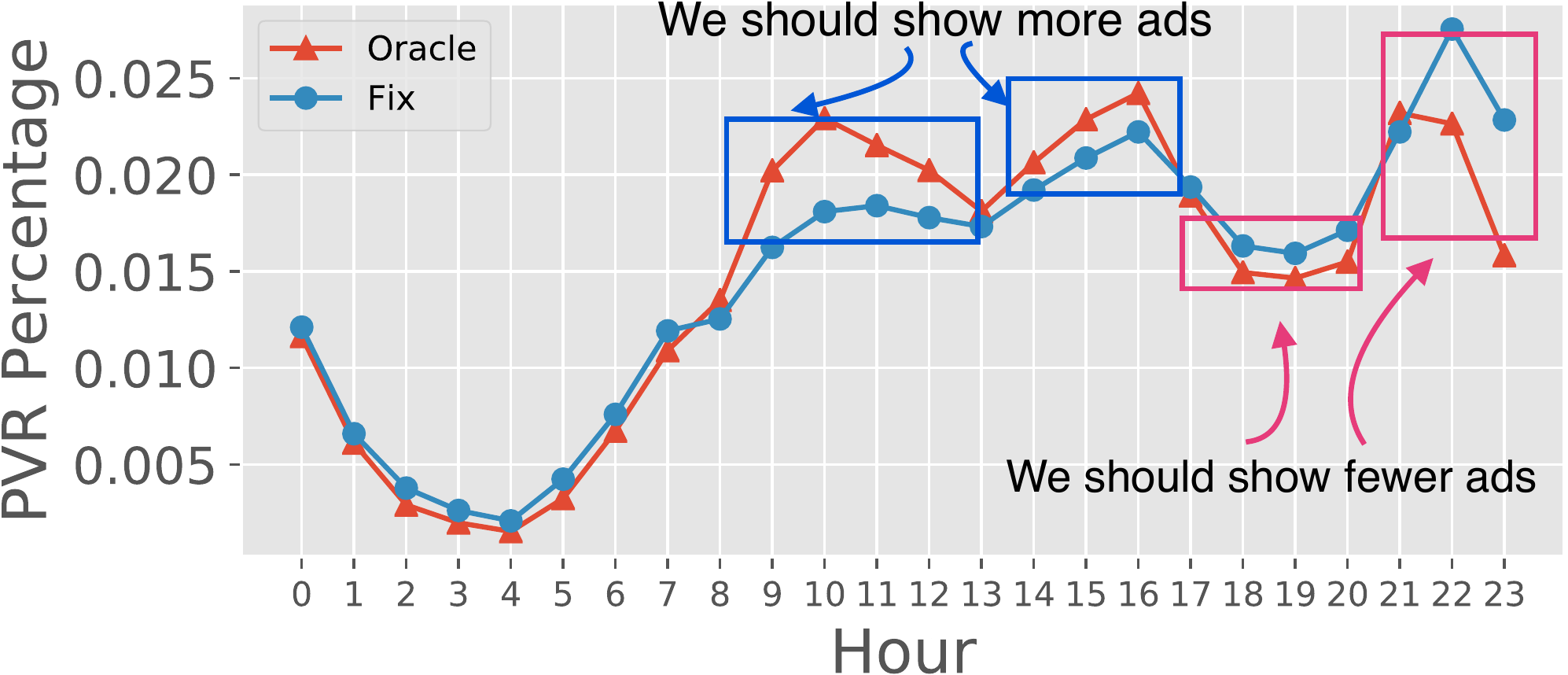}
 \par\vspace*{-8pt}%
\end{minipage}
\caption{The changing curves of per-hour advertising rates with one day. Fix: The proportion of ads exposed on each request is fixed and set to 0.35; Oracle: After one day, we could figure out the best dynamical advertising rate available for each hour under conditions that satisfy the daily constraint: $PV\hspace{-1pt}R=0.35$ through data analysis. }
\label{replay_a}
\end{figure}

\subsubsection*{Why higher level control can improve performance?}We analyze the finally exposed ads and all the corresponding candidate ads of each request within a day. First, we set the advertising rate of each request to a fixed value 0.35. Then we calculate out the proportion of the finally exposed ads within each hour to the total number of ads in that day, which is represented as the Fix policy in  Fig. \ref{replay_a}. Keeping the total number of ads displayed in a day exactly the same, Oracle is calculated by resorting all the candidates of all requests together according to the scores and picks out the top 35\% ads to display. Note that this Oracle policy shown in Fig. \ref{replay_a} is the best strategy available for displaying ads in one day. We can clearly find out that during the time period of hour 8 - hour 12, the advertising rate of the Oracle policy is more than 35\%, which means that we should display more ads within this period to enlarge revenue. Accordingly, during hour 17 - hour 20 and hour 22 - hour 23, the advertising rate of the Oracle policy is less than 35\%, which means that we should reduce the number of the unpromising ads and leave this opportunity to the more valuable ones. The detailed advertising performance of each hour is shown in Table. \ref{tabe3_per}. We can clearly see that the revenue gap between the baseline policy and our approaches mainly appear on hour 8 - hour 12; Besides, our method can obtain more cost-effective advertising exposure within hour 8 - hour 12 and hour 15 - hour 16. 
Our method can dynamically adjust the advertising number corresponding to different time periods with the daily PVR constraint satisfied.

\subsection{Online Results}
Lastly, we also report the production A/B test experiments, which compared the performance of our approach to a currently deployed baseline (which displays a fixed number of ads to every user \footnote{ In the online test, we have also tried the manual approach (as with the experimental setup). However, we found the manual method could not ensure a relatively stable satisfaction of the pvr constraint, so we omit the results for fair comparisons.}) in Taobao's online platform.
We conduct the experiments on the section of "guess what your like", where a mixture of recommendations and advertisements are displayed to the users. Our method does not fix the numbers and positions of ads. We apply our designed mechanism to adaptively adjust the scores of each ad to different users so as to display different numbers of ads to different users in different positions. More detials are illustrated in section \ref{AEM}.
For a fair comparison, we keep the \emph{platform-level} constraint the same for all approaches. As a result, we find that our approach does present different numbers of ads to different users in different positions satisfying the preset constraint overall. Besides, we observe 9\%, 3\%, and 2\% improvements in RPM (Revenue Per Mille), CTR (Click-through rate) and GMV (Gross Merchandise Value) respectively, which indicates that our adaptive exposure mechanism not only significantly increase the revenue of the platform (RPM) but also the revenues of advertisers (GMV). We also introduced the detailed online implementation process in the appendix.

\section{Conclusion} 
We first investigate the flaws in traditional E-commerce systems using fixed-positions to expose ads, and further propose more flexible advertising methods (Adaptive Exposure Mechanism) to alleviate the defects of fixed-positions. 
Further, we emphasize that there are a series of challenges when applying adaptive Exposure Mechanism in actual scenarios. 
We first model it as a psCMDP problem with different level constraints, and propose a constrained two-level reinforcement learning framework to solve this problem. 
Our framework offers high adaptability and quick response to the dynamic changing e-commerce environments. 
We also propose a novel replay buffer mechanism, CHER, to accelerate the policy training of the lower level. 
We have demonstrated that our designed adaptive Exposure Mechanism can provide more flexible advertising displaying methods while satisfying a series of constraints through offline simulation experiments and online verification. At the same time, we also verified that the constrained two-level reinforcement learning framework can effectively utilize the adaptive Exposure Mechanism to improve the platform revenue and user experience while satisfying the constraints.

\bibliographystyle{ACM-Reference-Format}
\bibliography{cikm}


\begin{thebibliography}{41}


\ifx \showCODEN    \undefined \def \showCODEN     #1{\unskip}     \fi
\ifx \showDOI      \undefined \def \showDOI       #1{#1}\fi
\ifx \showISBNx    \undefined \def \showISBNx     #1{\unskip}     \fi
\ifx \showISBNxiii \undefined \def \showISBNxiii  #1{\unskip}     \fi
\ifx \showISSN     \undefined \def \showISSN      #1{\unskip}     \fi
\ifx \showLCCN     \undefined \def \showLCCN      #1{\unskip}     \fi
\ifx \shownote     \undefined \def \shownote      #1{#1}          \fi
\ifx \showarticletitle \undefined \def \showarticletitle #1{#1}   \fi
\ifx \showURL      \undefined \def \showURL       {\relax}        \fi
\providecommand\bibfield[2]{#2}
\providecommand\bibinfo[2]{#2}
\providecommand\natexlab[1]{#1}
\providecommand\showeprint[2][]{arXiv:#2}

\bibitem[\protect\citeauthoryear{Achiam, Held, Tamar, and Abbeel}{Achiam
  et~al\mbox{.}}{2017}]%
        {achiam2017constrained}
\bibfield{author}{\bibinfo{person}{Joshua Achiam}, \bibinfo{person}{David
  Held}, \bibinfo{person}{Aviv Tamar}, {and} \bibinfo{person}{Pieter Abbeel}.}
  \bibinfo{year}{2017}\natexlab{}.
\newblock \showarticletitle{Constrained policy optimization}.
\newblock \bibinfo{journal}{\emph{arXiv preprint arXiv:1705.10528}}
  (\bibinfo{year}{2017}).
\newblock


\bibitem[\protect\citeauthoryear{Agrawal, Devanur, and Li}{Agrawal
  et~al\mbox{.}}{2016}]%
        {agrawal2016efficient}
\bibfield{author}{\bibinfo{person}{Shipra Agrawal}, \bibinfo{person}{Nikhil~R
  Devanur}, {and} \bibinfo{person}{Lihong Li}.}
  \bibinfo{year}{2016}\natexlab{}.
\newblock \showarticletitle{An efficient algorithm for contextual bandits with
  knapsacks, and an extension to concave objectives}. In
  \bibinfo{booktitle}{\emph{Proceedings of COLT}}. \bibinfo{pages}{4--18}.
\newblock


\bibitem[\protect\citeauthoryear{Altman}{Altman}{1999}]%
        {altman1999constrained}
\bibfield{author}{\bibinfo{person}{Eitan Altman}.}
  \bibinfo{year}{1999}\natexlab{}.
\newblock \bibinfo{booktitle}{\emph{Constrained Markov decision processes}}.
  Vol.~\bibinfo{volume}{7}.
\newblock \bibinfo{publisher}{CRC Press}.
\newblock


\bibitem[\protect\citeauthoryear{Ammar, Tutunov, and Eaton}{Ammar
  et~al\mbox{.}}{2015}]%
        {ammar2015safe}
\bibfield{author}{\bibinfo{person}{Haitham~Bou Ammar}, \bibinfo{person}{Rasul
  Tutunov}, {and} \bibinfo{person}{Eric Eaton}.}
  \bibinfo{year}{2015}\natexlab{}.
\newblock \showarticletitle{Safe policy search for lifelong reinforcement
  learning with sublinear regret}. In \bibinfo{booktitle}{\emph{Proceedings of
  ICML}}. \bibinfo{pages}{2361--2369}.
\newblock


\bibitem[\protect\citeauthoryear{Andrychowicz, Wolski, Ray, Schneider, Fong,
  Welinder, McGrew, Tobin, Abbeel, and Zaremba}{Andrychowicz
  et~al\mbox{.}}{2017}]%
        {andrychowicz2017hindsight}
\bibfield{author}{\bibinfo{person}{Marcin Andrychowicz}, \bibinfo{person}{Filip
  Wolski}, \bibinfo{person}{Alex Ray}, \bibinfo{person}{Jonas Schneider},
  \bibinfo{person}{Rachel Fong}, \bibinfo{person}{Peter Welinder},
  \bibinfo{person}{Bob McGrew}, \bibinfo{person}{Josh Tobin},
  \bibinfo{person}{OpenAI~Pieter Abbeel}, {and} \bibinfo{person}{Wojciech
  Zaremba}.} \bibinfo{year}{2017}\natexlab{}.
\newblock \showarticletitle{Hindsight experience replay}. In
  \bibinfo{booktitle}{\emph{Proceedings of NIPS}}. \bibinfo{pages}{5048--5058}.
\newblock


\bibitem[\protect\citeauthoryear{Bachrach, Ceppi, Kash, Key, and
  Kurokawa}{Bachrach et~al\mbox{.}}{2014}]%
        {bachrach2014optimising}
\bibfield{author}{\bibinfo{person}{Yoram Bachrach}, \bibinfo{person}{Sofia
  Ceppi}, \bibinfo{person}{Ian~A Kash}, \bibinfo{person}{Peter Key}, {and}
  \bibinfo{person}{David Kurokawa}.} \bibinfo{year}{2014}\natexlab{}.
\newblock \showarticletitle{Optimising trade-offs among stakeholders in ad
  auctions}. In \bibinfo{booktitle}{\emph{Proceedings of EC}}. ACM,
  \bibinfo{pages}{75--92}.
\newblock


\bibitem[\protect\citeauthoryear{Bacon, Harb, and Precup}{Bacon
  et~al\mbox{.}}{2017}]%
        {bacon2017option}
\bibfield{author}{\bibinfo{person}{Pierre-Luc Bacon}, \bibinfo{person}{Jean
  Harb}, {and} \bibinfo{person}{Doina Precup}.}
  \bibinfo{year}{2017}\natexlab{}.
\newblock \showarticletitle{The Option-Critic Architecture.}. In
  \bibinfo{booktitle}{\emph{Proceedings of AAAI}}. \bibinfo{pages}{1726--1734}.
\newblock


\bibitem[\protect\citeauthoryear{Badanidiyuru, Langford, and
  Slivkins}{Badanidiyuru et~al\mbox{.}}{2014}]%
        {badanidiyuru2014resourceful}
\bibfield{author}{\bibinfo{person}{Ashwinkumar Badanidiyuru},
  \bibinfo{person}{John Langford}, {and} \bibinfo{person}{Aleksandrs
  Slivkins}.} \bibinfo{year}{2014}\natexlab{}.
\newblock \showarticletitle{Resourceful contextual bandits}. In
  \bibinfo{booktitle}{\emph{Proceedings of COLT}}. \bibinfo{pages}{1109--1134}.
\newblock


\bibitem[\protect\citeauthoryear{Cai, Ren, Zhang, Malialis, Wang, Yu, and
  Guo}{Cai et~al\mbox{.}}{2017}]%
        {cai2017real}
\bibfield{author}{\bibinfo{person}{Han Cai}, \bibinfo{person}{Kan Ren},
  \bibinfo{person}{Weinan Zhang}, \bibinfo{person}{Kleanthis Malialis},
  \bibinfo{person}{Jun Wang}, \bibinfo{person}{Yong Yu}, {and}
  \bibinfo{person}{Defeng Guo}.} \bibinfo{year}{2017}\natexlab{}.
\newblock \showarticletitle{Real-time bidding by reinforcement learning in
  display advertising}. In \bibinfo{booktitle}{\emph{Proceedings of WSDM}}.
  ACM, \bibinfo{pages}{661--670}.
\newblock


\bibitem[\protect\citeauthoryear{Cai, Filos-Ratsikas, Tang, and Zhang}{Cai
  et~al\mbox{.}}{2018}]%
        {cai2018reinforcement1}
\bibfield{author}{\bibinfo{person}{Qingpeng Cai}, \bibinfo{person}{Aris
  Filos-Ratsikas}, \bibinfo{person}{Pingzhong Tang}, {and}
  \bibinfo{person}{Yiwei Zhang}.} \bibinfo{year}{2018}\natexlab{}.
\newblock \showarticletitle{Reinforcement Mechanism Design for e-commerce}. In
  \bibinfo{booktitle}{\emph{Proceedings of WWW}}. International World Wide Web
  Conferences Steering Committee, \bibinfo{pages}{1339--1348}.
\newblock


\bibitem[\protect\citeauthoryear{Chen, Yu, Da, Tan, Huang, and Tang}{Chen
  et~al\mbox{.}}{2018}]%
        {chen2018stabilizing}
\bibfield{author}{\bibinfo{person}{Shi-Yong Chen}, \bibinfo{person}{Yang Yu},
  \bibinfo{person}{Qing Da}, \bibinfo{person}{Jun Tan},
  \bibinfo{person}{Hai-Kuan Huang}, {and} \bibinfo{person}{Hai-Hong Tang}.}
  \bibinfo{year}{2018}\natexlab{}.
\newblock \showarticletitle{Stabilizing reinforcement learning in dynamic
  environment with application to online recommendation}. In
  \bibinfo{booktitle}{\emph{Proceedings of SIGKDD}}. ACM,
  \bibinfo{pages}{1187--1196}.
\newblock


\bibitem[\protect\citeauthoryear{Chow, Ghavamzadeh, Janson, and Pavone}{Chow
  et~al\mbox{.}}{2017}]%
        {chow2017risk}
\bibfield{author}{\bibinfo{person}{Yinlam Chow}, \bibinfo{person}{Mohammad
  Ghavamzadeh}, \bibinfo{person}{Lucas Janson}, {and} \bibinfo{person}{Marco
  Pavone}.} \bibinfo{year}{2017}\natexlab{}.
\newblock \showarticletitle{Risk-Constrained Reinforcement Learning with
  Percentile Risk Criteria}.
\newblock \bibinfo{journal}{\emph{Journal of Machine Learning Research}}
  \bibinfo{volume}{18} (\bibinfo{year}{2017}), \bibinfo{pages}{167--1}.
\newblock


\bibitem[\protect\citeauthoryear{Geibel}{Geibel}{2006}]%
        {geibel2006reinforcement}
\bibfield{author}{\bibinfo{person}{Peter Geibel}.}
  \bibinfo{year}{2006}\natexlab{}.
\newblock \showarticletitle{Reinforcement learning for MDPs with constraints}.
  In \bibinfo{booktitle}{\emph{Proceedings of ECML}}. Springer,
  \bibinfo{pages}{646--653}.
\newblock


\bibitem[\protect\citeauthoryear{Goodfellow, Bengio, Courville, and
  Bengio}{Goodfellow et~al\mbox{.}}{2016}]%
        {goodfellow2016deep}
\bibfield{author}{\bibinfo{person}{Ian Goodfellow}, \bibinfo{person}{Yoshua
  Bengio}, \bibinfo{person}{Aaron Courville}, {and} \bibinfo{person}{Yoshua
  Bengio}.} \bibinfo{year}{2016}\natexlab{}.
\newblock \bibinfo{booktitle}{\emph{Deep learning}}. Vol.~\bibinfo{volume}{1}.
\newblock \bibinfo{publisher}{MIT press Cambridge}.
\newblock


\bibitem[\protect\citeauthoryear{Hester, Vecerik, Pietquin, Lanctot, Schaul,
  Piot, Horgan, Quan, Sendonaris, Osband, et~al\mbox{.}}{Hester
  et~al\mbox{.}}{2018}]%
        {hester2018deep}
\bibfield{author}{\bibinfo{person}{Todd Hester}, \bibinfo{person}{Matej
  Vecerik}, \bibinfo{person}{Olivier Pietquin}, \bibinfo{person}{Marc Lanctot},
  \bibinfo{person}{Tom Schaul}, \bibinfo{person}{Bilal Piot},
  \bibinfo{person}{Dan Horgan}, \bibinfo{person}{John Quan},
  \bibinfo{person}{Andrew Sendonaris}, \bibinfo{person}{Ian Osband},
  {et~al\mbox{.}}} \bibinfo{year}{2018}\natexlab{}.
\newblock \showarticletitle{Deep q-learning from demonstrations}. In
  \bibinfo{booktitle}{\emph{Proceedings of AAAI}}.
\newblock


\bibitem[\protect\citeauthoryear{Hu, Da, Zeng, Yu, and Xu}{Hu
  et~al\mbox{.}}{2018}]%
        {hu2018reinforcement}
\bibfield{author}{\bibinfo{person}{Yujing Hu}, \bibinfo{person}{Qing Da},
  \bibinfo{person}{Anxiang Zeng}, \bibinfo{person}{Yang Yu}, {and}
  \bibinfo{person}{Yinghui Xu}.} \bibinfo{year}{2018}\natexlab{}.
\newblock \showarticletitle{Reinforcement Learning to Rank in E-Commerce Search
  Engine: Formalization, Analysis, and Application}. In
  \bibinfo{booktitle}{\emph{Proceedings of SIGKDD}}.
\newblock


\bibitem[\protect\citeauthoryear{Ieong, Mahdian, and Vassilvitskii}{Ieong
  et~al\mbox{.}}{2014}]%
        {ieong2014advertising}
\bibfield{author}{\bibinfo{person}{Samuel Ieong}, \bibinfo{person}{Mohammad
  Mahdian}, {and} \bibinfo{person}{Sergei Vassilvitskii}.}
  \bibinfo{year}{2014}\natexlab{}.
\newblock \showarticletitle{Advertising in a stream}. In
  \bibinfo{booktitle}{\emph{Proceedings of WWW}}. ACM, \bibinfo{pages}{29--38}.
\newblock


\bibitem[\protect\citeauthoryear{Jaderberg, Mnih, Czarnecki, Schaul, Leibo,
  Silver, and Kavukcuoglu}{Jaderberg et~al\mbox{.}}{2016}]%
        {jaderberg2016reinforcement}
\bibfield{author}{\bibinfo{person}{Max Jaderberg}, \bibinfo{person}{Volodymyr
  Mnih}, \bibinfo{person}{Wojciech~Marian Czarnecki}, \bibinfo{person}{Tom
  Schaul}, \bibinfo{person}{Joel~Z Leibo}, \bibinfo{person}{David Silver},
  {and} \bibinfo{person}{Koray Kavukcuoglu}.} \bibinfo{year}{2016}\natexlab{}.
\newblock \showarticletitle{Reinforcement learning with unsupervised auxiliary
  tasks}.
\newblock \bibinfo{journal}{\emph{arXiv preprint arXiv:1611.05397}}
  (\bibinfo{year}{2016}).
\newblock


\bibitem[\protect\citeauthoryear{Jin, Song, Li, Gai, Wang, and Zhang}{Jin
  et~al\mbox{.}}{2018}]%
        {jin2018real}
\bibfield{author}{\bibinfo{person}{Junqi Jin}, \bibinfo{person}{Chengru Song},
  \bibinfo{person}{Han Li}, \bibinfo{person}{Kun Gai}, \bibinfo{person}{Jun
  Wang}, {and} \bibinfo{person}{Weinan Zhang}.}
  \bibinfo{year}{2018}\natexlab{}.
\newblock \showarticletitle{Real-Time Bidding with Multi-Agent Reinforcement
  Learning in Display Advertising}. In \bibinfo{booktitle}{\emph{Proceedings of
  CIKM}}. ACM, \bibinfo{pages}{2193--2201}.
\newblock


\bibitem[\protect\citeauthoryear{Kulkarni, Narasimhan, Saeedi, and
  Tenenbaum}{Kulkarni et~al\mbox{.}}{2016}]%
        {kulkarni2016hierarchical}
\bibfield{author}{\bibinfo{person}{Tejas~D Kulkarni}, \bibinfo{person}{Karthik
  Narasimhan}, \bibinfo{person}{Ardavan Saeedi}, {and} \bibinfo{person}{Josh
  Tenenbaum}.} \bibinfo{year}{2016}\natexlab{}.
\newblock \showarticletitle{Hierarchical deep reinforcement learning:
  Integrating temporal abstraction and intrinsic motivation}. In
  \bibinfo{booktitle}{\emph{Proceedings of NIPS}}. \bibinfo{pages}{3675--3683}.
\newblock


\bibitem[\protect\citeauthoryear{LeCun, Bengio, and Hinton}{LeCun
  et~al\mbox{.}}{2015}]%
        {lecun2015deep}
\bibfield{author}{\bibinfo{person}{Yann LeCun}, \bibinfo{person}{Yoshua
  Bengio}, {and} \bibinfo{person}{Geoffrey Hinton}.}
  \bibinfo{year}{2015}\natexlab{}.
\newblock \showarticletitle{Deep learning}.
\newblock \bibinfo{journal}{\emph{Nature}} \bibinfo{volume}{521},
  \bibinfo{number}{7553} (\bibinfo{year}{2015}), \bibinfo{pages}{436}.
\newblock


\bibitem[\protect\citeauthoryear{Lee, Orten, Dasdan, and Li}{Lee
  et~al\mbox{.}}{2012}]%
        {lee2012estimating}
\bibfield{author}{\bibinfo{person}{Kuang-chih Lee}, \bibinfo{person}{Burkay
  Orten}, \bibinfo{person}{Ali Dasdan}, {and} \bibinfo{person}{Wentong Li}.}
  \bibinfo{year}{2012}\natexlab{}.
\newblock \showarticletitle{Estimating conversion rate in display advertising
  from past erformance data}. In \bibinfo{booktitle}{\emph{Proceedings of
  SIGKDD}}. ACM, \bibinfo{pages}{768--776}.
\newblock


\bibitem[\protect\citeauthoryear{Levine, Finn, Darrell, and Abbeel}{Levine
  et~al\mbox{.}}{2016}]%
        {levine2016end}
\bibfield{author}{\bibinfo{person}{Sergey Levine}, \bibinfo{person}{Chelsea
  Finn}, \bibinfo{person}{Trevor Darrell}, {and} \bibinfo{person}{Pieter
  Abbeel}.} \bibinfo{year}{2016}\natexlab{}.
\newblock \showarticletitle{End-to-end training of deep visuomotor policies}.
\newblock \bibinfo{journal}{\emph{JMLR}} \bibinfo{volume}{17},
  \bibinfo{number}{1} (\bibinfo{year}{2016}), \bibinfo{pages}{1334--1373}.
\newblock


\bibitem[\protect\citeauthoryear{Lillicrap, Hunt, Pritzel, Heess, Erez, Tassa,
  Silver, and Wierstra}{Lillicrap et~al\mbox{.}}{2015}]%
        {lillicrap2015continuous}
\bibfield{author}{\bibinfo{person}{Timothy~P Lillicrap},
  \bibinfo{person}{Jonathan~J Hunt}, \bibinfo{person}{Alexander Pritzel},
  \bibinfo{person}{Nicolas Heess}, \bibinfo{person}{Tom Erez},
  \bibinfo{person}{Yuval Tassa}, \bibinfo{person}{David Silver}, {and}
  \bibinfo{person}{Daan Wierstra}.} \bibinfo{year}{2015}\natexlab{}.
\newblock \showarticletitle{Continuous control with deep reinforcement
  learning}.
\newblock \bibinfo{journal}{\emph{arXiv preprint arXiv:1509.02971}}
  (\bibinfo{year}{2015}).
\newblock


\bibitem[\protect\citeauthoryear{McMahan, Holt, Sculley, Young, Ebner, Grady,
  Nie, Phillips, Davydov, Golovin, et~al\mbox{.}}{McMahan
  et~al\mbox{.}}{2013}]%
        {mcmahan2013ad}
\bibfield{author}{\bibinfo{person}{H~Brendan McMahan}, \bibinfo{person}{Gary
  Holt}, \bibinfo{person}{David Sculley}, \bibinfo{person}{Michael Young},
  \bibinfo{person}{Dietmar Ebner}, \bibinfo{person}{Julian Grady},
  \bibinfo{person}{Lan Nie}, \bibinfo{person}{Todd Phillips},
  \bibinfo{person}{Eugene Davydov}, \bibinfo{person}{Daniel Golovin},
  {et~al\mbox{.}}} \bibinfo{year}{2013}\natexlab{}.
\newblock \showarticletitle{Ad click prediction: a view from the trenches}. In
  \bibinfo{booktitle}{\emph{Proceedings of SIGKDD}}. ACM,
  \bibinfo{pages}{1222--1230}.
\newblock


\bibitem[\protect\citeauthoryear{Mehta et~al\mbox{.}}{Mehta
  et~al\mbox{.}}{2013}]%
        {mehta2013online}
\bibfield{author}{\bibinfo{person}{Aranyak Mehta} {et~al\mbox{.}}}
  \bibinfo{year}{2013}\natexlab{}.
\newblock \showarticletitle{Online matching and ad allocation}.
\newblock \bibinfo{journal}{\emph{Foundations and Trends in Theoretical
  Computer Science}} \bibinfo{volume}{8}, \bibinfo{number}{4}
  (\bibinfo{year}{2013}), \bibinfo{pages}{265--368}.
\newblock


\bibitem[\protect\citeauthoryear{Mnih, Kavukcuoglu, Silver, Rusu, Veness,
  Bellemare, Graves, Riedmiller, Fidjeland, Ostrovski, et~al\mbox{.}}{Mnih
  et~al\mbox{.}}{2015}]%
        {mnih2015human}
\bibfield{author}{\bibinfo{person}{Volodymyr Mnih}, \bibinfo{person}{Koray
  Kavukcuoglu}, \bibinfo{person}{David Silver}, \bibinfo{person}{Andrei~A
  Rusu}, \bibinfo{person}{Joel Veness}, \bibinfo{person}{Marc~G Bellemare},
  \bibinfo{person}{Alex Graves}, \bibinfo{person}{Martin Riedmiller},
  \bibinfo{person}{Andreas~K Fidjeland}, \bibinfo{person}{Georg Ostrovski},
  {et~al\mbox{.}}} \bibinfo{year}{2015}\natexlab{}.
\newblock \showarticletitle{Human-level control through deep reinforcement
  learning}.
\newblock \bibinfo{journal}{\emph{Nature}} \bibinfo{volume}{518},
  \bibinfo{number}{7540} (\bibinfo{year}{2015}), \bibinfo{pages}{529}.
\newblock


\bibitem[\protect\citeauthoryear{Perlich, Dalessandro, Hook, Stitelman, Raeder,
  and Provost}{Perlich et~al\mbox{.}}{2012}]%
        {perlich2012bid}
\bibfield{author}{\bibinfo{person}{Claudia Perlich}, \bibinfo{person}{Brian
  Dalessandro}, \bibinfo{person}{Rod Hook}, \bibinfo{person}{Ori Stitelman},
  \bibinfo{person}{Troy Raeder}, {and} \bibinfo{person}{Foster Provost}.}
  \bibinfo{year}{2012}\natexlab{}.
\newblock \showarticletitle{Bid optimizing and inventory scoring in targeted
  online advertising}. In \bibinfo{booktitle}{\emph{Proceedings of SIGKDD}}.
  ACM, \bibinfo{pages}{804--812}.
\newblock


\bibitem[\protect\citeauthoryear{Prashanth and Ghavamzadeh}{Prashanth and
  Ghavamzadeh}{2016}]%
        {prashanth2016variance}
\bibfield{author}{\bibinfo{person}{LA Prashanth} {and}
  \bibinfo{person}{Mohammad Ghavamzadeh}.} \bibinfo{year}{2016}\natexlab{}.
\newblock \showarticletitle{Variance-constrained actor-critic algorithms for
  discounted and average reward MDPs}.
\newblock \bibinfo{journal}{\emph{Machine Learning}} \bibinfo{volume}{105},
  \bibinfo{number}{3} (\bibinfo{year}{2016}), \bibinfo{pages}{367--417}.
\newblock


\bibitem[\protect\citeauthoryear{Schulman, Levine, Abbeel, Jordan, and
  Moritz}{Schulman et~al\mbox{.}}{2015a}]%
        {schulman2015trust}
\bibfield{author}{\bibinfo{person}{John Schulman}, \bibinfo{person}{Sergey
  Levine}, \bibinfo{person}{Pieter Abbeel}, \bibinfo{person}{Michael Jordan},
  {and} \bibinfo{person}{Philipp Moritz}.} \bibinfo{year}{2015}\natexlab{a}.
\newblock \showarticletitle{Trust region policy optimization}. In
  \bibinfo{booktitle}{\emph{Proceedings of ICML}}. \bibinfo{pages}{1889--1897}.
\newblock


\bibitem[\protect\citeauthoryear{Schulman, Moritz, Levine, Jordan, and
  Abbeel}{Schulman et~al\mbox{.}}{2015b}]%
        {schulman2015high}
\bibfield{author}{\bibinfo{person}{John Schulman}, \bibinfo{person}{Philipp
  Moritz}, \bibinfo{person}{Sergey Levine}, \bibinfo{person}{Michael Jordan},
  {and} \bibinfo{person}{Pieter Abbeel}.} \bibinfo{year}{2015}\natexlab{b}.
\newblock \showarticletitle{High-dimensional continuous control using
  generalized advantage estimation}.
\newblock \bibinfo{journal}{\emph{arXiv preprint arXiv:1506.02438}}
  (\bibinfo{year}{2015}).
\newblock


\bibitem[\protect\citeauthoryear{Sutton, Barto, et~al\mbox{.}}{Sutton
  et~al\mbox{.}}{1998}]%
        {sutton1998reinforcement}
\bibfield{author}{\bibinfo{person}{Richard~S Sutton}, \bibinfo{person}{Andrew~G
  Barto}, {et~al\mbox{.}}} \bibinfo{year}{1998}\natexlab{}.
\newblock \bibinfo{booktitle}{\emph{Reinforcement learning: An introduction}}.
\newblock \bibinfo{publisher}{MIT press}.
\newblock


\bibitem[\protect\citeauthoryear{Tang, Rosales, Singh, and Agarwal}{Tang
  et~al\mbox{.}}{2013}]%
        {tang2013automatic}
\bibfield{author}{\bibinfo{person}{Liang Tang}, \bibinfo{person}{Romer
  Rosales}, \bibinfo{person}{Ajit Singh}, {and} \bibinfo{person}{Deepak
  Agarwal}.} \bibinfo{year}{2013}\natexlab{}.
\newblock \showarticletitle{Automatic ad format selection via contextual
  bandits}. In \bibinfo{booktitle}{\emph{Proceedings of CIKM}}. ACM,
  \bibinfo{pages}{1587--1594}.
\newblock


\bibitem[\protect\citeauthoryear{Tessler, Mankowitz, and Mannor}{Tessler
  et~al\mbox{.}}{2018}]%
        {tessler2018reward}
\bibfield{author}{\bibinfo{person}{Chen Tessler}, \bibinfo{person}{Daniel~J
  Mankowitz}, {and} \bibinfo{person}{Shie Mannor}.}
  \bibinfo{year}{2018}\natexlab{}.
\newblock \showarticletitle{Reward Constrained Policy Optimization}.
\newblock \bibinfo{journal}{\emph{arXiv preprint arXiv:1805.11074}}
  (\bibinfo{year}{2018}).
\newblock


\bibitem[\protect\citeauthoryear{Uchibe and Doya}{Uchibe and Doya}{2007}]%
        {uchibe2007constrained}
\bibfield{author}{\bibinfo{person}{Eiji Uchibe} {and} \bibinfo{person}{Kenji
  Doya}.} \bibinfo{year}{2007}\natexlab{}.
\newblock \showarticletitle{Constrained reinforcement learning from intrinsic
  and extrinsic rewards}. In \bibinfo{booktitle}{\emph{Proceedings of ICDL}}.
  IEEE, \bibinfo{pages}{163--168}.
\newblock


\bibitem[\protect\citeauthoryear{Wang, Zhang, and Yuan}{Wang
  et~al\mbox{.}}{2016}]%
        {wang2016display}
\bibfield{author}{\bibinfo{person}{Jun Wang}, \bibinfo{person}{Weinan Zhang},
  {and} \bibinfo{person}{Shuai Yuan}.} \bibinfo{year}{2016}\natexlab{}.
\newblock \showarticletitle{Display advertising with real-time bidding (RTB)
  and behavioural targeting}.
\newblock \bibinfo{journal}{\emph{arXiv preprint arXiv:1610.03013}}
  (\bibinfo{year}{2016}).
\newblock


\bibitem[\protect\citeauthoryear{Wu, Chen, Yang, Wang, Tan, Zhang, and Gai}{Wu
  et~al\mbox{.}}{2018}]%
        {wu2018budget}
\bibfield{author}{\bibinfo{person}{Di Wu}, \bibinfo{person}{Xiujun Chen},
  \bibinfo{person}{Xun Yang}, \bibinfo{person}{Hao Wang}, \bibinfo{person}{Qing
  Tan}, \bibinfo{person}{Xiaoxun Zhang}, {and} \bibinfo{person}{Kun Gai}.}
  \bibinfo{year}{2018}\natexlab{}.
\newblock \showarticletitle{Budget Constrained Bidding by Model-free
  Reinforcement Learning in Display Advertising}. In
  \bibinfo{booktitle}{\emph{Proceedings of CIKM}}. ACM,
  \bibinfo{pages}{1443--1451}.
\newblock


\bibitem[\protect\citeauthoryear{Wu, Srikant, Liu, and Jiang}{Wu
  et~al\mbox{.}}{2015}]%
        {wu2015algorithms}
\bibfield{author}{\bibinfo{person}{Huasen Wu}, \bibinfo{person}{R Srikant},
  \bibinfo{person}{Xin Liu}, {and} \bibinfo{person}{Chong Jiang}.}
  \bibinfo{year}{2015}\natexlab{}.
\newblock \showarticletitle{Algorithms with logarithmic or sublinear regret for
  constrained contextual bandits}. In \bibinfo{booktitle}{\emph{Proceedings of
  NIPS}}. \bibinfo{pages}{433--441}.
\newblock


\bibitem[\protect\citeauthoryear{Zhang, Yuan, and Wang}{Zhang
  et~al\mbox{.}}{2014}]%
        {zhang2014optimal}
\bibfield{author}{\bibinfo{person}{Weinan Zhang}, \bibinfo{person}{Shuai Yuan},
  {and} \bibinfo{person}{Jun Wang}.} \bibinfo{year}{2014}\natexlab{}.
\newblock \showarticletitle{Optimal real-time bidding for display advertising}.
  In \bibinfo{booktitle}{\emph{Proceedings of SIGKDD}}. ACM,
  \bibinfo{pages}{1077--1086}.
\newblock


\bibitem[\protect\citeauthoryear{Zhao, Qiu, Guan, Zhao, and He}{Zhao
  et~al\mbox{.}}{2018b}]%
        {zhao2018deep}
\bibfield{author}{\bibinfo{person}{Jun Zhao}, \bibinfo{person}{Guang Qiu},
  \bibinfo{person}{Ziyu Guan}, \bibinfo{person}{Wei Zhao}, {and}
  \bibinfo{person}{Xiaofei He}.} \bibinfo{year}{2018}\natexlab{b}.
\newblock \showarticletitle{Deep Reinforcement Learning for Sponsored Search
  Real-time Bidding}.
\newblock \bibinfo{journal}{\emph{Proceedings of SIGKDD}}
  (\bibinfo{year}{2018}).
\newblock


\bibitem[\protect\citeauthoryear{Zhao, Li, An, Lu, Yang, and Chu}{Zhao
  et~al\mbox{.}}{2018a}]%
        {zhao2018impression}
\bibfield{author}{\bibinfo{person}{Mengchen Zhao}, \bibinfo{person}{Zhao Li},
  \bibinfo{person}{Bo An}, \bibinfo{person}{Haifeng Lu}, \bibinfo{person}{Yifan
  Yang}, {and} \bibinfo{person}{Chen Chu}.} \bibinfo{year}{2018}\natexlab{a}.
\newblock \showarticletitle{Impression Allocation for Combating Fraud in
  E-commerce Via Deep Reinforcement Learning with Action Norm Penalty}. In
  \bibinfo{booktitle}{\emph{Proceedings of IJCAI}}.
  \bibinfo{pages}{3940--3946}.
\newblock


\end{thebibliography}

\appendix
\section{Appendix}

\subsection{Discussion: Adaptive Exposure}
Current research on dynamic ad exposure focuses on sponsored search\cite{bachrach2014optimising}, stream advertising in news feeds\cite{ieong2014advertising}, etc. Their dynamic ad exposure mainly refers to how to select the appropriate location and quantity to display the ad in the fixed optional ad positions\cite{bachrach2014optimising}, or how to dynamically insert the ad in the feeds through the user's previous browsing process\cite{ieong2014advertising}. Compared with sponsored search\cite{bachrach2014optimising}, our mechanism does not limit the position of advertisements. Instead, it chooses the number and location of ads by means of score sorting, which means that our approach can bring more flexibility. Compared with stream advertising in news feeds\cite{ieong2014advertising} we consider the mixed-display scenario where both recommended products and advertised products are displayed altogether to the customers and their display orders are determined by their relative rankings. 

\subsection{Related Work}
\subsubsection{Bidding Optimization in Real-Time Bidding}
Under the Real-Time Bidding (RTB) settings in E-commerce advertising, amounts of work have been proposed to estimate the impression values, e.g. click-through rate (CTR) \cite{mcmahan2013ad} and conversion rate (CVR) \cite{lee2012estimating}, which help to improve the bidding effectiveness via predicting more precise impression values. 
Besides, the user impression analysis, bidding optimization is one of another most concerned problems in RTB, whose goal is to dynamically set a more appropriate price for each auction aiming at maximizing some key performance indicators (KPIs) (e.g. CTR) \cite{wang2016display}. However, constraints are inevitable while solving optimization problems in real world bidding situations. So, smarter bidding strategies are needed for attaining higher KPI values (e.g. the cumulative impression value), which can be achieved through reinforcement learning techniques \cite{perlich2012bid,zhang2014optimal,cai2017real}. In these approaches, they optimize the bidding strategy under the fixed budget constraint and the budget will be reset at the beginning of each episode.
Perlich et al. \cite{perlich2012bid} and Zhang et al. \cite{zhang2014optimal} propose static bid optimization frameworks based on the distribution analysis of the previously collected log data. However, their approach can't apply well to the setting in which the data distribution is unstable and will change from day to day in extreme circumstances. For this reason, Cai et al. \cite{cai2017real} model the bidding problem as a MDP and consider the budgets allocation as a sequential decision problem. Experimental results show the robustness of their reinforcement learning approach under the non-stationary auction environments. 

By contrast, we are the first to propose a more general deep reinforcement learning framework which takes more realistic business constraints into consideration. In our settings, we concentrate on the practical \textit{advertising with adaptive exposure problem}.
Not only do we consider the trajectory level constraint $C_T$, but also the state level constraint $C_S$. This is the main reason why the previous approaches are not applicable to our settings.

\subsubsection{Constrained Reinforcement Learning}
We are focusing on a constrained optimization problem and a number of researches have been done. One typical solution is the constrained reinforcement learning. \citet{uchibe2007constrained} propose a policy gradient algorithm which uses gradient projection to enforce the active constraints. However, their approach is unable to prevent the policy from becoming insecure at the beginning of the training. Later, \citet{ammar2015safe} propose a theoretically-motivated policy gradient method for lifelong learning under safety constraints. Unfortunately, they involve an expensive inner loop which contains an optimization of a semi-definite program, making it unsuitable for the DRL settings.
Similarly, \citet{chow2017risk} propose a primal-dual sub-gradient method for risk-constrained reinforcement learning, which takes policy gradient steps trading off return for lower risk while simultaneously learning the trade-off coefficients (dual variables).

More recently, a number of DRL-based approaches have been proposed to address the constrained optimization problem. \citet{achiam2017constrained} use the conjugate gradient method to optimize the policy. However, the computational cost will significantly arise as the constraint number increases, resulting in such approaches inapplicable. \citet{tessler2018reward} propose the Reward Constrained Policy Optimization (RCPO), which converts the trajectory constraints into per-state penalties and dynamically adjusts the weight of each per-state penalty during the learning procedure via propagating the constraint violation signal over the entire trajectory.

Our work tackle the multi-constraint problem from a different point of view and take the relationship between the different constraints into account. We decouple the original multi-constraints optimization problem into relatively independent single constraint optimization problems and propose a constrained two-level reinforcement learning framework. More importantly, our two-level framework is quite general and any state-of-the-art RL algorithms can be flexibly applied to learning procedures of both levels.

\subsection{Network structure and training parameters}
\subsubsection{CHER}
Both the actor network and the critic network are four-layer fully connected neural networks, where each of the two hidden layers consists of 20 neurons and a ReLU activation function is applied on the outputs of the hidden layers. A tanh function is applied to the output layer of the actor network to bound the size of the adjusted scores. The input of the actor network and critic network is a tensor of shape 46 representative feature vectors of the request's candidate ad items and the number of currently exposed items. The output of the actor network and the critic network are respectively 15 actions and corresponding Q-values. The learning rate of the actor is 0.001, the learning rate of the critic is 0.0001, and the size of the replay buffer is 50000. The exploration rate starts from 1 and decays linearly to 0.001 after 50,000 steps.It is worth pointing out that in the environment, we will make certain adjustments to the action, such as adding a certain value, performing certain scaling, to ensure that the operation of adjusting the score is in line with the business logic. Therefore, the output of the action is not the actual adjusted scores. We consider this adjustment part as part of the environmental logic. It does not affect the training of the network.

\subsubsection{Higher Level Control}
DQN network has three-layer neural networks. The hidden layer consists of 20 neurons and a ReLU activation function is applied on the outputs of the hidden layer. Then we connect the hidden layer output to: 1) the nodes with the same number of actions, which is used to simulate the action advantage value $A$, 2) only one node, which is used to simulate the state value $V$. Finally we obtain $Q(s,a) = V(S) + (A(s, a)- \frac{1}{|A|}\sum_{a'} A(s, a'))$. The size of the replay buffer is 5000. We use the prioritized replay to sample the replay buffer. The learning rate is 0.0006. The exploration rate starts from 1 and linearly decays to 0.001 after 1000 steps. Also we set the discount factor $\gamma=1$.

\subsection{Experimental Setup}
Based on the log data, for each request, 
we collect 15 recommended products and their scores marked by the recommendation system as candidates for each request, and the information of 15 advertising products, such as: eCPM (effective cost per mille), price, predicted Click-Through-Rate (pCTR), and initial score. Since the actual amount of data is significantly large, we sample a part of the data for empirical evaluation, and verify that the sampled data has representativeness to the real data set. In both training and evaluation stages, we split the previously collected data by day and replay the requests in chronological order for simulation. We consider the data flow from 00:00 AM to the next day as a trajectory. At the beginning of each day, the number of ads has been displayed and number of requests have been reset to 0. At the end of each day, we count the daily number of ads displayed and make a judgement whether the trajectory-level constraint $C_T$ has been satisfied.

\begin{figure}
\centering
\begin{minipage}[b]{0.8\linewidth}
 \includegraphics[width=1\textwidth]{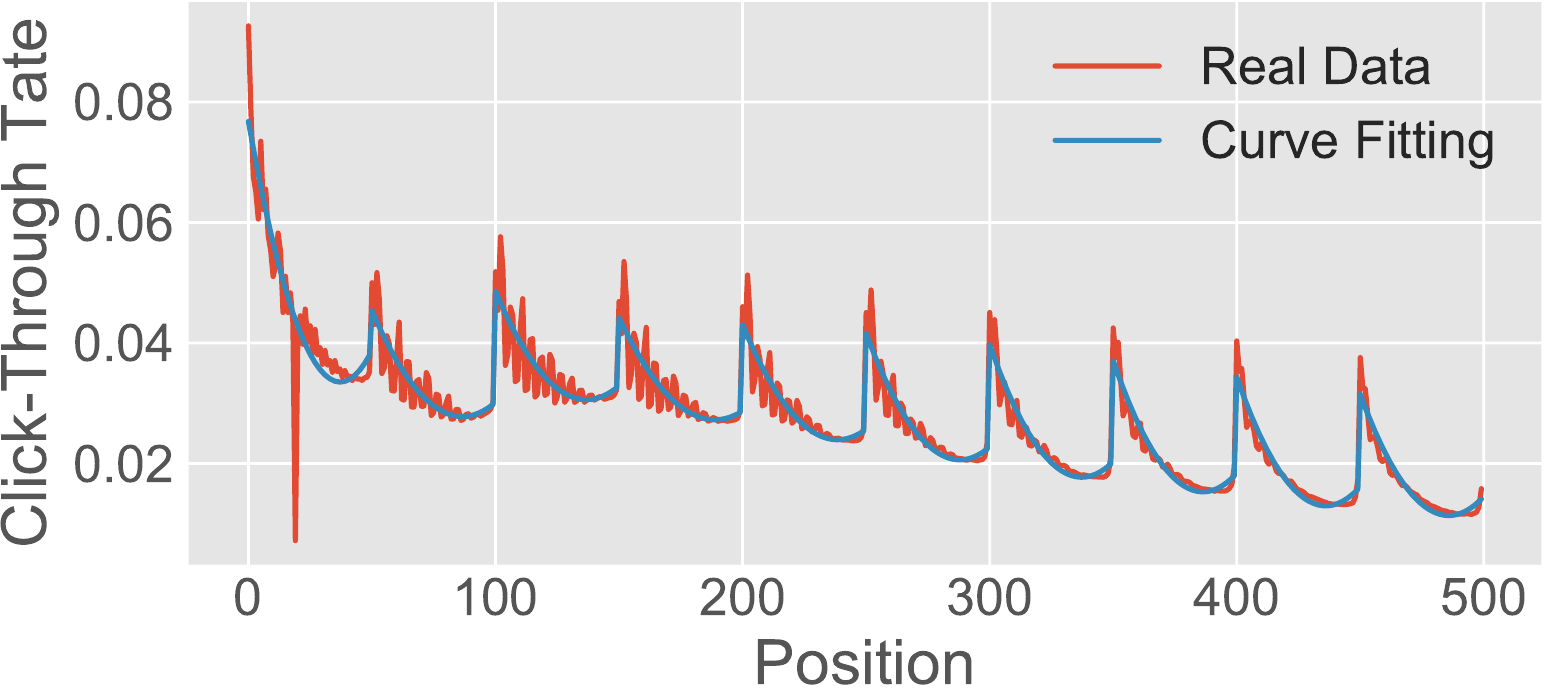}
 \par\vspace*{-8pt}%
\end{minipage}%
\caption{The impact of different positions on CTR.}
\label{replay_b}
\end{figure}

A state consists of 46 dimensions including the characteristics of the 15 candidate ads: eCPM, price, pCTR, and the number of exposed ads. Action is the coefficient to adjust scores for 15 ads, $action = \{\eta_1, \eta_2, \dots, \eta_{15}\}$. After adjusting scores using actions, 15 candidate advertising commodities and 15 candidate recommended commodities are sorted based on the new scores. The reward is calculated according to the ads in the first 10 exposure items.
We can replay the data to train and test the effect of our algorithm offline in two ways: 1) after ads adjusting scores, whether the quantity of ads in the 10 exposure items meets $C_T$ and $C_{S}$, and 2) the rewards of the exposed ads. Actually, the positions of ads have impact on user behaviors. E.g., the ads in front are more possible to be clicked, and so on. Hence the reward is defined as: 
\begin{equation}
r = \sum_{d} f_{p}(d)\ \times\ eCPM(d) 
\end{equation}
where $eCPM(d)$ is the eCPM value of the ad $d$, and $f_{p}(d)$ 
corrects the eCPM by considering the influence of different positions. $f_{p}(d)$ is fitted using the real data. (Fig. \ref{replay_b})

\subsection{Online Experiment}
Due to the architecture of the online engine in Taobao, we replaced the gradient based DDPG with the widely used gradient-free Cross Entropy Method (CEM, an evolution based genetic algorithm) within the platform. When deploying our algorithm to the online environment, we consider two separate processes: (1) online serving and data collection; (2) offline training. For (1), we use Blink (an open source stream processing framework which is specially designed and optimized for e-commerce scenarios) to record the constantly updated online data. Besides, to fully explore the parameter space of CEM, we split the online traffic into a number of buckets and deploy different sets of parameter configurations at the same time. Different buckets are controlled by different parameters. After processing each user's request, the newly produced data is recorded to corresponding data tables by Blink. For (2), a centralized learner will periodically update its parameters based on the latest recorded data, generate different sets of parameters for different buckets and synchronize them to the parameter server. At the same time, the online search engine deployed at each bucket also regularly request the latest parameters from the parameter server.

\end{document}